%% file: paper-icra21-mc.tex
\documentclass[letterpaper, 10 pt, conference]{ieeeconf}
\pdfoutput=1

\IEEEoverridecommandlockouts
\makeatletter
\let\NAT@parse\undefined
\makeatother

\usepackage{mathrsfs}
\usepackage{amsmath,amssymb,amsfonts}
\usepackage{mathtools,bbm,bm}
\usepackage{pgfplots}
\usepackage{tikz}
\usepackage{algorithm}
\usepackage[noend]{algpseudocode}
\usepackage{textcomp}
\usepackage{xcolor}
\usepackage[numbers,sort&compress]{natbib}
\usepackage{subcaption}
\usepackage{makeidx}
\usepackage{graphicx}
\usepackage{multicol}
\usepackage[bottom]{footmisc}
\usepackage{array}
\usepackage{balance}
\usepackage{comment}
\usepackage{xparse}

\usepackage{amsthm}

\usepackage[
hidelinks,
colorlinks,
linkcolor=black,
citecolor=black,
filecolor=red,
urlcolor=blue
]{hyperref}
\usepackage{url}

\usepackage[binary-units=true]{siunitx}
\usepackage{pgffor}
\usepackage{dsfont}

\usepackage{pbox} % Adaptive width paragraph boxes for latex
\usepackage{tabularx} % Centered cells and better table stretching
\usepackage{booktabs} %for \multicolumn{n}{c}{text}, also top/bottomrules

\usepackage{pgfplotstable} % fancy tables from csv
\usepackage{pgfplots}
\usepackage{tikz}

\usepgfplotslibrary{external}
\pgfplotsset{compat=newest} % enable magic compatibility
\pgfplotsset{plot coordinates/math parser=false}

\DeclareGraphicsExtensions{.pdf,.png,.jpg}

\newlength\figureheight
\newlength\figurewidth
\newlength\subfiguresheight
\newlength\phantomheight
\newlength\compositewidth

\newlength\notationsymbol
\newlength\notationdescription

\renewcommand\thesubsubsection{%
  \thesubsection\alph{subsubsection}%
}

\newif\ifshowdevelopment

\showdevelopmentfalse

\newif\ifextended
\extendedtrue

\graphicspath{{./fig/}}

\newcommand{\belowtoprule}{\midrule[\heavyrulewidth]}

\newtheorem{theorem}{Theorem}%[section]
\algnewcommand{\CommentLine}[1]{\State \(\triangleright\) #1}

\DeclareMathOperator*{\argmax}{arg\,max}

\newcommand{\spacefont}[1]{{\mathbb{#1}}}

\newcommand{\real}{{\spacefont{R}}}

\renewcommand{\vec}[1]{\mathbf{#1}}

\newcommand{\probabilityfont}[1]{{\mathds{#1}}}
\newcommand{\E}{{\probabilityfont{E}}}

\renewcommand{\H}{{\probabilityfont{H}}} % entropy
\newcommand{\MI}{{\probabilityfont{I}}}

\newcommand{\setsystemfont}[1]{{\mathscr{#1}}}
\newcommand{\ground}{\setsystemfont{U}}
\newcommand{\independence}{\setsystemfont{I}}

\newcommand{\block}{\setsystemfont{B}}

\newcommand{\setfun}{g} % we use g to avoid confusion with dynamics

\newcommand{\opt}{{\star}}

\newcommand{\state}{\vec{x}}
\newcommand{\states}{\vec{X}}
\newcommand{\observation}{\vec{y}}
\newcommand{\observations}{\vec{Y}}
\newcommand{\environment}{E}

\newcommand{\sensingquality}{J}
\newcommand{\quota}{B}
\newcommand{\control}{u}
\newcommand{\dynamics}{f}
\newcommand{\observationfunction}{h}

\newcommand{\camera}{{F^\mathrm{cam}}}

\newcommand{\safespace}{\mathcal{X}_{\text{safe}}}

\newcommand{\viewthreshold}{\epsilon_{\text{view}}}

\newcommand{\distancefactor}{\alpha}
\newcommand{\environmentsguess}{\mathcal{E}_{\text{dist}}}
\newcommand{\cell}{c}
\newcommand{\covercells}{C_{\text{cov}}}
\newcommand{\cellweight}{w_{\text{cell}}}
\newcommand{\newcellweight}{w_{\text{new}}} % new cells

\newcommand{\robots}{{\mathcal{R}}}
\newcommand{\cells}{{\mathcal{C}}}
\newcommand{\numrobot}{{n_\mathrm{r}}}

\newcommand{\numcells}{{n_\mathrm{m}}}
\newcommand{\numrounds}{{n_\mathrm{d}}}

\newcommand{\controls}{\mathcal{U}}

\newcommand{\optionalsubscript}[1][]{%
  \ifthenelse{\equal{#1}{}}{%
  }{%
    $_{#1}$%
  }%
}

\NewDocumentCommand{\distalgnamed}{O{} O{} m}{%
  \ifthenelse{\equal{#2}{}}{%
    {\text{\textnormal{#3}\optionalsubscript[#1]{}}}%
  }{%
    {\text{#2{#3}\optionalsubscript[#1]{}}}%
  }%
}

\NewDocumentCommand{\dgreedy}{O{} O{}}
{%
  \distalgnamed[#1][#2]{DSGA}%
}

\NewDocumentCommand{\rsp}{O{} O{}}
{%
  {\distalgnamed[#1][#2]{RSP}}%
}
\NewDocumentCommand{\rrsp}{O{} O{}}
{%
  {\distalgnamed[#1][#2]{R-lRSP}}%
}

\newcommand{\inputfigure}[2][]
{%
  \ifthenelse{\equal{#1}{}}%
  {}{\tikzsetnextfilename{#1}}%
  \input{#2}%
}

\newcommand{\numrobots}{{\numrobot}}

\newcommand{\viewvalue}{\setfun_{\text{view}}}
\newcommand{\distancevalue}{\setfun_{\text{dist}}}
\newcommand{\expectedcoverage}{\setfun_{\text{cov}}}
\newcommand{\actionstostates}{\Phi}

\begin{document}

\title{\LARGE
  Volumetric Objectives for Multi-Robot Exploration of Three-Dimensional
  Environments
}

\author{%
Micah Corah
and
Nathan Michael
\thanks{
  Micah Corah is affiliated with the NASA Jet Propulsion Laboratory, California
  Institute of Technology.
  Nathan Michael is affiliated with the Robotics Institute, Carnegie Mellon
  University (CMU).
  This work was completed while Micah Corah was a student at CMU.
}
\thanks{
  \tt{micah.d.corah@jpl.nasa.gov, nmichael@cmu.edu}%
}
\thanks{This work was supported by industry.}%
}

\maketitle

\begin{abstract}
  Volumetric objectives for exploration and perception tasks seek to capture a
  sense of value (or reward) for hypothetical observations at one or more
  camera views for robots operating in unknown environments.
  For example, a volumetric objective may reward robots proportionally to the
  expected volume of unknown space to be observed.
  We identify connections between existing information-theoretic and
  coverage objectives in terms of expected coverage,
  particularly that mutual information without noise is a special case of
  expected coverage.
  Likewise, we provide the first comparison, of which we are aware, between
  information-based approximations and coverage objectives for exploration,
  and we find, perhaps surprisingly, that coverage objectives can
  significantly outperform information-based objectives in practice.
  Additionally, the analysis for information and coverage objectives
  demonstrates that Randomized Sequential Partitions---a method for efficient
  distributed sensor planning---applies for both classes of objectives,
  and we provide simulation results in a variety of environments for as many as
  32 robots.
\end{abstract}

\section{Introduction}

Efficient exploration of unstructured three-dimensional environments can enable
mapping of caves~\citep{tabib2020} and assistance in search and
rescue~\citep{murphy2016} and is key to autonomy for aerial and ground
robots.
Planners for such robots seek to maximize various measures of reward for
future camera views such as the volume of unknown space that the robots will
observe.
However, when the environment geometry is unknown, there is uncertainty
in what the robots will observe and, in turn, how to assign rewards to as-yet
unknown observations.

Methods for assigning volumetric rewards can be divided roughly into two groups:
coverage-like and information-theoretic methods.
Coverage methods compute rewards for observations given an assumed realization
of the environment~\citep{%
simmons2000aaai,butzke2011planning,delmerico2018auro,bircher2018receding}.
While these approaches are simple the rewards they compute may not be
representative of the actual observations.
Alternatively, information-theoretic methods seek to compute reward (mutual
information) given a distribution over environments via Bayesian
priors~\citep{%
julian2014,charrow2015icra,charrow2015rss,zhang2019ijrr,henderson2020icra}.
However, information-theoretic methods are limited:
all assume that cells in the map are independent;
and, while rewards for depth observations along individual rays are accurate,
rewards for multiple camera views are approximate~\citep{charrow2015icra}.
This work relates these two families via connections to \emph{expected coverage},
and our results, the first we are aware of comparing these two
classes, suggest that coverage objectives may outperform existing
information-theoretic methods in practice.

\begin{comment}
First, environments for exploration do not reflect the common assumption that
cells are occupied with independent probability.
Such cell independence assumptions are central to existing
information-theoretic
objectives~\citep{charrow2015icra,julian2014,zhang2019ijrr}
which often strongly emphasize efficient
computation~\citep{charrow2015icra,zhang2019ijrr,li2019rss,henderson2020icra}.
Although these independence assumptions can be limiting, they admit changes
to the occupancy prior for unobserved space
(typically decreasing the prior)%
~\citep{henderson2019thesis,henderson2020icra,tabib2016iros,corah2019auro}.
Still, even though recent works~\citep{zhang2019ijrr,li2019rss}
establish that Shannon mutual information can be evaluated efficiently
and accurately for an individual range observation along a ray,
methods for evaluating the joint contributions of multiple rays and camera views
depend on approximation by summing over rays~\citep{charrow2015icra}.
Our analysis highlights connections between mutual information and coverage
objectives and suggests that coverage can accurately represent joint
contributions when the prior probability of occupancy is low.
Moreover, our results demonstrate that switching to a coverage objective can
improve completion times by as much as 16\%.
\end{comment}

The same connections between mutual information and coverage also enable
efficient distributed sensor planning.
Many multi-robot sensing problems can be solved efficiently via greedy
sequential planning methods for certain submodular
objectives~\citep{fisher1978,singh2009,atanasov2015icra,schlotfeldt2018ral}.
However, these algorithms require robots to plan sequentially which causes
planning time to increase with the size of the team and prevents designers from
taking advantage of parallel computation.
Some recent works study similar greedy planners where some robots can plan
in
parallel~\citep{corah2017rss,corah2018cdc,corah2020phd,grimsman2018tcns,gharesifard2017,sun2020acc}.
However, many of these works provide suboptimality guarantees that degrade with
increasing numbers of
robots~\citep{grimsman2018tcns,gharesifard2017,sun2020acc}.
Now, while our earlier work on distributed exploration did not provide a-priori
guarantees on
solution quality~\citep{corah2017rss}, our recent work on planning via
Randomized Sequential Partitions (\rsp{}) provides guarantees on
suboptimality that depend on an amount of
inter-robot redundancy and for a slightly narrower class of
objectives~\citep{corah2020phd,corah2018cdc}.
The analysis in this paper demonstrates that these suboptimality
guarantees are also applicable to exploration, for both mutual information and
coverage objectives.
Further, the results evaluate a variety of environments and numbers of robots
and demonstrate that planning via \rsp{} improves solution quality compared to
more myopic planners while running with fewer steps than existing sequential
methods.

Additionally, we note that an earlier (and slightly longer) version of this work
appears in~\citep[Chapter~7]{corah2020phd}.

\section{Background on submodular maximization}

This work applies methods for maximizing submodular set functions to solve
multi-robot receding-horizon sensor planning problems.
Consider a collection of \emph{finite} spaces of actions (or trajectories)
$\block_1,\ldots,\block_\numrobots$ called \emph{blocks} available to a team
of $\numrobots$ robots.
The set of all of these actions
$\ground = \bigcup_{r \in \{1, \ldots, \numrobots\}} \block_r$
is the \emph{ground set}, and, while planning, each robot can select one
action from its set of actions.
Further, the set of complete and partial assignments
$\independence = \{ I \subseteq \ground \mid  1 \geq |I\cap \block_r|\}$
forms a \emph{simple partition matroid}~\citep[Sec.~39.4]{schrijver2003}.

Then, a \emph{set function} $\setfun : 2^\ground\rightarrow\real$ maps a
collection of such actions to a scalar such as to quantify sensor coverage or
information gain.
We treat unions implicitly so that
$\setfun(A,B) = \setfun(A \cup B)$ for $A,B\subseteq\ground$
and automatically wrap elements of subsets in sets
so $\setfun(x) = \setfun(\{x\})$ for $x\in\ground$.

Monotonicity conditions are important to many works that seek maximize set
functions~\citep{nemhauser1978best,fisher1978}, and many sensing objectives are
known to have two such monotonicity conditions: monotonicity and
submodularity~\citep{krause2005uai}.
\citet{foldes2005} defined a sequence of such monotonicity conditions.
If a function is monotonically increasing, then for any
$X\subseteq\ground$ and $x \in \ground \setminus X$
marginal gains are always positive
\begin{align}
  \setfun(x|X) \geq 0
  \label{eq:monotonicity}
\end{align}
where $\setfun(x|X) = \setfun(x,X) - \setfun(X)$, and
this forms the 1-increasing condition~\citep{foldes2005}.
Further, submodularity states that marginal gains decrease in the presence of
a growing set of prior selections which forms the 2-decreasing condition.
Specifically, given
$A \! \subseteq \! B \subseteq \! \ground$ and
$x \! \in \! \ground \! \setminus \! B$
submodularity states that
\begin{align}
  \setfun(x|A) - \setfun(x|B) \leq 0.
\end{align}
The coverage and mutual information objectives we study are known
to be monotonic, submodular and normalized
($\setfun(\emptyset) = 0$).
Additionally, set functions may also be 3-increasing%
\footnote{%
  Note that \citep{corah2018cdc} refers to the 3-increasing condition as
  \emph{submodularity of conditioning}
  while \citep{corah2020phd} provides more detail on monotonicity conditions.
}
which is useful for modeling redundancy between pairs of
actions for efficient distributed planning~\citep{corah2020phd,corah2018cdc}.
While weighted coverage objectives are 3-increasing, submodular mutual
information objectives may not satisfy this requirement~\citep{corah2018cdc}.
We then seek to determine whether or not mutual information objectives
\emph{specific to exploration} are 3-increasing as well.

\section{Minimum time exploration problem}
\label{sec:exploration_problem}

Consider a team of robots $\robots=\{1,\ldots,\numrobot\}$ mapping a
discretized environment $\environment=[\cell_1,\ldots,\cell_{\numcells}]$
consisting of
cells $\cells = \{1,\ldots,\numcells\}$
that are each either \emph{free} ($\bm{0}$) or
\emph{occupied}
($\bm{1}$)
($\cell_i \in \{0,1\}$),
according to a probability distribution $\environmentsguess$.
Each robot $r\in\robots$ moves about with states
$\state_{t,r}\in\real^3$
at each \emph{discrete} time $t$
according to the dynamics
\begin{align}
  \state_{t,r} &= \dynamics(\state_{t-1,r}, \control_{t,r})
                     \label{eq:exploration_dynamics}
\end{align}
where $\control_{t,r}\in\controls$ belongs to a finite set of control inputs.
Robots must also remain in free space which induces the constraint that they
remain in a safe set
\begin{align}
  \state_{t,r} \in \safespace(\environment).
  \label{eq:safe_space}
\end{align}
As they move about, the robots observe the environment with depth cameras.
%(or other ray-based sensors such as lidar sensors).
Assuming measurements are deterministic, without noise,
robots can infer definitively that cells in the path of each ray are free up to
the first occupied cell or the maximum range.
Thus, when robots obtain observations from their depth cameras they can infer
the occupancy of the set of cells
$\camera(\state, \environment) \subseteq \cells$
which have occupancy values
\begin{align}
  \observation_{t,r} &=
  \observationfunction(\state_{t,r}, \environment) =
  \{(i, \cell_i) : i \in \camera(\state, \environment)\}.
  \label{eq:exploration_observation}
\end{align}

The robots seek to observe as much of the environment as possible in terms of
\emph{environment coverage}:
\begin{align}
  \sensingquality(\states_{1:t}, \observations_{1:t}) =
  \left|
  \bigcup\nolimits_{t'\in\{1,\ldots,t\},r\in\robots}
  \camera(\state_{t',r}, \environment)
  \right|.
  \label{eq:environment_coverage}
\end{align}
Specifically, we seek to \emph{minimize time to reach a threshold on
environment coverage}, completing the exploration task.
\begin{comment}
Exploration is complete once the environment coverage reaches an
environment-dependent quota $\quota(\environment)$ when
\begin{align}
  \sensingquality(\states_{1:T},\observations_{1:T})
  \geq \quota(\environment),
\end{align}
and the robots
seek to minimize the amount of time $T$ required to reach that quota and
complete the exploration process.
\end{comment}

\section{Planning for exploration}
\label{sec:receding-horizon_planning}

Robots explore via receding-horizon planning
and collectively maximize an objective $\setfun$ over an $L$-step horizon
\begin{align}
  \max_{\control'_{1:L,1:\numrobot}}~& \setfun(\states_{t:t+L,1:\numrobot})
  \nonumber \\
  \mathrm{s.t.}~
              & \state_{t+l,r} \in \safespace(\observations_{1:t})
  \nonumber \\
              & \state_{t+l,r} = \dynamics(\state_{t+l,r}, \control'_{l,r})
  \nonumber \\
              & \observation_{t+l,r} = \observationfunction(\state_{t+l,r},
                                                            \environment)
  \nonumber \\
              &\mbox{for all } l \in \{1\ldots L\} \mbox{ and }
              r \in \robots,
              \label{eq:receding_horizon_exploration}
\end{align}
where $\safespace(\observations_{1:t})$ refers to the set of states
\emph{known} to be safe given available observations
(unlike \eqref{eq:safe_space} which refers to the complete set of safe states).
Robots then execute the first control actions in the sequences
$\control'_{1,1:\numrobot}$ and repeat.

Additionally, we can interpret $\setfun$ as a set function and rewrite
\eqref{eq:receding_horizon_exploration}
as a submodular maximization problem with a simple partition matroid
constraint~\citep{corah2017rss,atanasov2015icra,schlotfeldt2018ral}.
From this perspective, the ground set consists of assignments of control actions
to robots $\ground = \robots \times \controls^L$,
and the blocks of the partition matroid
are assignments $\block_r = \{r\}\!\times \controls^L$ to robots
$r \in \robots$.

We propose an objective that consists of two components $\viewvalue$ and
$\distancevalue$ so that, given $X \subseteq \ground$, then
\begin{align}
  \setfun(X) &=
  \viewvalue(X)
  +
  \sum_{r \in \robots}
  \distancevalue(X_r),
  \label{eq:exploration_objective}
\end{align}
where $X_r = \block_r \cap X$ is the assignment to robot $r\in\robots$.
The \emph{view} (or volumetric) reward $\viewvalue$ seeks to capture the value
of observations from camera views over the planning horizon while the
\emph{distance} component $\distancevalue$ rewards moving toward
regions of the environment---often beyond the planning horizon---where valuable
observations can be obtained.

The distance reward is based on our prior work~\citep{corah2019ral} and
rewards robots for reducing the distance to the nearest view with value
($\viewvalue$) above a given threshold.
%(see Fig.~\ref{fig:view_distance}).
This can be thought of as an analogue of nearest frontier
exploration~\citep{yamauchi1997} for three-dimensional environments.
\begin{comment}
\begin{figure}
  \centering
  \includegraphics[width=0.8\linewidth]{view_distance}
  \caption[Visualization of sampled informative views and view distance]{%
    This visualization illustrates the distance reward.
    Sampled informative views in (red arrows) are analogous to frontiers
    (views being near the boundary of the unobserved space)
    and
    produce a goal region which robots should remain near.
    Robots then seek to minimize the distance to the nearest view
    (rainbow with red corresponding to least distance) during planning via the
    distance reward.
  }\label{fig:view_distance}
\end{figure}
\end{comment}
Note that $\setfun$ will remain normalized ($\setfun(\emptyset)=0$) and
retain the
monotonicity properties of $\viewvalue$ given that
$\distancevalue$ is non-negative and zero when trajectories are not assigned
(due to $\distancevalue$ being additive).%
%\footnote{%
  %Being additive, marginal gains for the distance reward are fixed and do not
  %depend on other robots.
  %As such, aside from increasing monotonically, \emph{all} higher-order
  %monotonicity conditions hold because \emph{all} second and higher-order
  %derivatives are zero.
%}

\subsection{Sequential greedy assignment for submodular maximization}
\label{sec:sequential_planning}

Sequentially maximizing $\setfun$ for each robot provides solutions to
receding-horizon planning problems
\eqref{eq:receding_horizon_exploration} within half
of optimal if individual robots plan optimally~\citep{fisher1978} or nearly
so if robots plan approximately~\citep{singh2009}
so long as objective is submodular, monotonic, and normalized.
Specifically, robots obtain solutions
$X^\mathrm{g} = \{x_1^\mathrm{g},\ldots,x_\numrobots^\mathrm{g}\}$
by planning greedily in sequence, maximizing $\setfun$ conditional
on prior solutions:
\begin{align}
  x_r^\mathrm{g} \in \argmax_{x \in \block_r} \setfun(x | X^\mathrm{g}_{1:r-1}).
  \label{eq:sequential_planning}
\end{align}
This approach is simple and relatively efficient given that obtaining optimal
solutions is NP-Hard.
However, distributed implementations are intractable for large teams of
robots as planning time increases because robots must plan in sequence rather
than in parallel.

\subsection{Distributed planning via Randomized Sequential Partitions}
\label{sec:rsp_planning}

If $\setfun$ is also 3-increasing, planning with Randomized Sequential
Partitions (\rsp{}) can take advantage of parallel computation and guarantee
solutions \emph{approaching the original bound of one-half}, depending on a
measure of redundancy between robots~\citep{corah2018cdc,corah2020phd}.

\rsp{} planners run in a fixed number of steps $\numrounds$ that does not
grow with the number of robots ($\numrobots$).
Robots planning via \rsp{} assign themselves independently and randomly to one
of $\numrounds$ sequential steps.
Then, robots assigned to the same step plan in parallel and maximize $\setfun$
conditional on decisions from prior planning steps.
Additionally, while this work does not focus on implementation,
readers may refer to \citet{corah2020phd} for a distributed, time-synchronous
implementation of \rsp{}.

\section{Volumetric rewards}

The volumetric reward $\viewvalue$ seeks to capture the joint value of
observations (camera views) in terms of the amount of unknown space that the
robots will collectively observe.
Ideally, $\viewvalue$ would correspond directly to actual increments in
environment coverage \eqref{eq:environment_coverage}.
Except, the environment $\environment$ is unknown and, in turn, so are future
values of the environment coverage.
To mitigate this issue, prior works predominantly either compute rewards
given an assumed~\citep{%
simmons2000aaai,butzke2011planning,delmerico2018auro,bircher2018receding}
(or possibly learned)
guess at the
environment instantiation
or else by approximating mutual information
for observations given a Bayesian prior%
~\citep{charrow2015icra,charrow2015rss,julian2014,henderson2020icra}
so far \emph{always} assuming
independent (Bernoulli) cell occupancy.
The following discussion unifies and contrasts these two families of exploration
rewards and demonstrates that both coverage and information-based rewards
satisfy the requirements for \rsp{} planning.

\subsection{Expected coverage}
\label{sec:expected_coverage_exploration}

We will relate different exploration rewards in terms of expected coverage.
Consider some assumed distribution over possible environments
$\environmentsguess$
with the requirement that
any environment $\environmentsguess$ assigns non-zero probability
$E'\sim\environmentsguess$ must be consistent with observations up to the
current time $\observations_{1:t}$
(although $\environmentsguess$ may assign non-zero probability to only a single
realization).
For convenience, define the set of \emph{future} states that robots will
visit while executing actions $X\subseteq\ground$ as
$\actionstostates(X)$.
Given non-negative rewards for observing each cell
$\cellweight : \cells \rightarrow \real_{\geq 0}$
the expected weighted coverage is
\begin{align}
  \expectedcoverage(X) &=
  \E_{\environment'\sim\environmentsguess}\left[
    \sum_{i \in \covercells(X,\environment')} \cellweight(i)
  \right]
  \label{eq:expected_coverage}
\end{align}
where $\covercells$ is the set of cells that the robots could observe:
\begin{align}
  \covercells(X,\environment')=
  % new observations
  \bigcup\nolimits_{\state\in\Phi(X)}\camera(\state,\environment').
  \label{eq:covercells}
\end{align}
We apply a weighting wcheme which
provides a unit reward for each newly observed cell
\begin{align}
  \newcellweight(i) &=
  \begin{cases}
    % already observed observation
    0 &
      i \in
      \bigcup\nolimits_{t'\in\{1,\ldots,t\},r\in\robots}
      \camera(\state_{t',r}, \environment)
    \\
    % new observation
    1 & \text{otherwise (newly observed cells)}
  \end{cases}.
  \label{eq:uniform_weighting}
\end{align}
However, other weighting schemes are also relevant.
For example, \citet{yoder2016fsr} propose an objective for inspection tasks
that encourages observing surfaces.
%Their surface frontier approach would be similar in spirit to a scheme that
%provides increased weight to unobserved cells that are near known occupied
%cells.
Later, we will also find weighting cells by entropy produces a mutual
information objective.
%\footnote{
%  If not for mutual information, we would define the weighted expected coverage
%  \eqref{eq:expected_coverage}, concisely, in terms of unobserved cells and the
%  weight \eqref{eq:uniform_weighting} as a constant for all cells.
%  However, probabilistic occupancy does not lend itself to explicit distinctions
%  between known and unknown values.
%}

\subsubsection{Expected coverage retains monotonicity properties}
\label{sec:expected_coverage_monotonicity}

The expected coverage \eqref{eq:expected_coverage} is normalized, monotonic,
submodular, and 3-increasing
because these conditions hold for weighted
coverage~\citep[Theorem~5]{corah2018cdc}
and because the expectation over environments in \eqref{eq:expected_coverage}
forms a convex combination which preserves monotonicity conditions
(and being normalized)~\citep{foldes2005}.
Likewise, suboptimality guarantees for \rsp{}
planning~\citep{corah2018cdc,corah2020phd}
apply to receding-horizon planning for exploration with expected coverage.

\subsection{Noiseless mutual information for depth sensors}
\label{sec:mutual_information_exploration}

Most existing works on mutual information for occupancy mapping assume noisy
measurements via either a general~\citep{julian2014} noise model or simplified
(often Gaussian) models~\citep{charrow2015icra,zhang2019ijrr}.
However, \citet{zhang2019ijrr} observed that the choice of information
metric and noise model has little impact on performance in exploration
experiments.
Likewise, \citet{henderson2020icra} noted that sensor
noise for modern lidar sensors and depth cameras is typically small compared to
the maximum range.
For these reasons, we assume that sensor noise is negligible for the purpose of
evaluating mutual information for
exploration\footnote{%
  Alternatively, sensor noise may be more relevant to perception
  tasks such as surface reconstruction.
}
and so ignore sensor noise in this work.
Additionally, we assume that
cell occupancy probabilities are independent according to the prior $\environmentsguess$
like previous works on mutual information for mapping%
~\citep{charrow2015icra,julian2014,zhang2019ijrr}.

The combination of cell independence and lack of sensor noise produces a special
case of \emph{expected coverage}:

\begin{theorem}[%
  Noiseless mutual information with independent cells is
  expected coverage]
  \label{theorem:mapping_information_is_coverage}
  The mutual information\footnote{
    \citet{cover2012} provide more detail on mutual information and entropy.
  }
  $\MI(\environment; \observations(X))$
  between an environment $\environment$
  with uncertain occupancy according to the distribution $\environmentsguess$
  and hypothetical future observations $\observations(X)$ can be written as
  \begin{align}
    \MI(\environment; \observations(X)) &=
    \E_{\environment' \sim \environmentsguess}
    \left[
      \sum\nolimits_{i \in \covercells(X, \environment')} \H(\cell_i)
    \right].
    \label{eq:mapping_information_independent}
  \end{align}
  where $\H(\cell_i)$ is the entropy of cell $\cell_i$, assuming:
  \begin{enumerate}
    \item Cell occupancy probabilities $\environmentsguess$ are independent, and
    \item There is no sensor noise.
  \end{enumerate}
  This expression \eqref{eq:mapping_information_independent}
  has the form of expected weighted coverage%
  \footnote{
    Entropy $\H$ is always non-negative~\citep{cover2012} which satisfies
    the requirements in Sec.~\ref{sec:expected_coverage_exploration}.
  }
  \eqref{eq:expected_coverage}
  and is therefore 3-increasing.
\end{theorem}

The proof is included in
\ifextended
Appendix~\ref{appendix:proof_that_mapping_information_is_coverage}.
\else
Appendix~I of the extended version~\citep{corah2021icraarxiv}.
\fi
Theorem~\ref{theorem:mapping_information_is_coverage} produces insights into
objective design and implies
that suboptimality guarantees for $\rsp$ planning for 3-increasing functions
apply to noiseless
mutual information just as for expected coverage, even though mutual information
is not 3-increasing in general~\citep{corah2018cdc}.

\subsection{Design of volumetric objectives for exploration}
\label{sec:objectives_remarks}

Let us now expand on the prior two sections
(Sec.~\ref{sec:expected_coverage_exploration} and
Sec.~\ref{sec:mutual_information_exploration})
which define volumetric exploration objectives and their
properties and discuss the ramifications of the choice of a volumetric reward.

\subsubsection{Optimistic coverage}
\label{sec:optimistic_coverage}

The definition of expected coverage also applies to degenerate priors over
environments.
\emph{Optimistic coverage} is an expected coverage objective
that \emph{optimistically} assumes that unobserved space is empty and provides
unit rewards~\eqref{eq:uniform_weighting} on unobserved cells; and
numerous works on exploration develop similar
objectives~\citep{simmons2000aaai,butzke2011planning}.
Moreover, evaluation of optimistic coverage consists simply of counting newly
observed cells.
Thus, implementation is trivial even for joint observations by teams of robots.

\subsubsection{Occupancy priors for mutual information and optimism}
\label{sec:occupancy_priors}

Mapping applications and works
exploration~\citep{julian2014,charrow2015icra,zhang2019ijrr}
frequently assume that unobserved cells are independent and occupied with a
probability of 0.5.
However, the environments that robots explore may predominantly consist
of open space, and robots may actually observe more unknown space than they
would otherwise predict.
For this reason, our prior works~\citep{tabib2016iros,corah2019auro} select
priors with relatively low occupancy probabilities.
Similarly, \citet{henderson2019thesis} provides detailed
discussion and illustrations that demonstrate how the occupancy prior can
affect decisions.

A prior with a low occupancy probability assumes that unobserved
cells are unoccupied like the optimistic coverage objective.
As such, \emph{optimistic coverage is equivalent to mutual information}
in the limit after applying a scaling factor to
\eqref{eq:mapping_information_independent}
to normalize entropies of
unobserved cells,\footnote{%
  The entropy of unobserved cells is constant and depends on the prior.
}
assuming entropies of observed cells are zero.
This is useful because optimistic coverage for multiple robots can be evaluated
exactly while methods based on mutual information are approximate for multiple
camera views~\citep{charrow2015icra,henderson2020icra,zhang2019ijrr}.

\section{Online bounds on solution quality}
\label{sec:online_bounds_exploration}

The 3-increasing condition establishes that \rsp{} satisfies
problem-dependent suboptimality guarantees for solving
\eqref{eq:receding_horizon_exploration}.
Additionally, submodular objectives, such as the ones in this work, satisfy
certain online bounds~\citep{minoux1978} which can provide much tighter
guarantees for \emph{specific solutions}%
~\citep{leskovec2007,golovin2011jair,krause2008}.
Further, these bounds apply to \emph{any feasible solution} which makes them
suitable for comparison across different planners---even planners
that do not satisfy specific guarantees.

While existing works study these online bounds in the context of
cardinality-constrained problems, adapting these bounds to
simple partition matroids is straightforward.
Consider any---possibly incomplete---feasible solution $X\in\independence$
to a receding-horizon planning problem as in
Sec.~\ref{sec:receding-horizon_planning}
and an optimal solution $X^\opt$.
The following holds, respectively, by monotonicity, submodularity, and greedy
choice:
\begin{align}
  \setfun(X^\opt)
  &\leq \setfun(X,X^\opt)
  \leq \setfun(X) + \sum_{x^\opt \in X^\opt} \setfun(x^\opt | X)
  \nonumber\\
  &\leq \setfun(X) + \sum_{r \in \robots} \max_{x\in\block_r} \setfun(x | X).
  \label{eq:online_bound}
\end{align}
We apply two instances of the above bound.
In the first, the \emph{online} bound, $X$ is the full solution returned by the
planner (assigning actions to all robots).
Next, we call the case where $X=\emptyset$ (the empty set) the \emph{oblivious}
bound.\footnote{%
  Because \eqref{eq:online_bound} provides an upper bound on $\setfun(X^\opt)$
  we can obtain a lower bound on suboptimality by computing the
  \emph{ratio of the solution value to the right-hand-side}.
}
%This oblivious bound reduces to
%$\setfun(X^\opt) \leq \sum_{r \in \robots} \max_{x\in\block_r} \setfun(x)$
%which would produce an optimal solution if $\setfun$ were additive (modular).

\begin{figure}
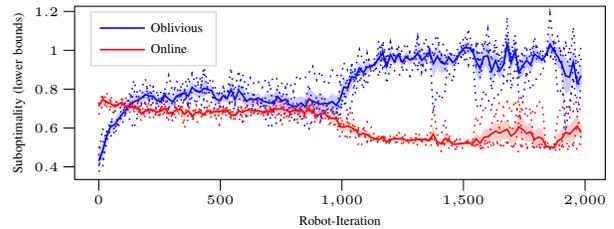

  \centering
  \setlength{\figurewidth}{\linewidth}
  \setlength{\figureheight}{0.45\linewidth}
  \tiny
  \inputfigure[online_and_oblivious_bounds]{%
  ./fig/test_standard_configuration_suboptimality_boxes_0b8bea2.tex}
  \caption[Illustration of online and oblivious solution bounds]{
    The above illustrates the online and oblivious
    bounds on suboptimality (solution value over bound) with five trials,
    sequential planning, and 16 robots with robots starting near the \emph{same
    position}.
    The online bound is tighter early when robots are close together,
    and the oblivious bound becomes tighter later, as robots spread out.
    Note that
    the exploration process is nearly complete by 1000 robot-iterations, and
    evaluation is not exact due to approximation of
    \eqref{eq:online_bound} with MCTS.
    \emph{Shaded regions here and in the rest of the results delineate the
    standard error.}
  }%
  \label{fig:online_bounds_example}
\end{figure}

Figure~\ref{fig:online_bounds_example} illustrates these bounds for a small set
of exploration experiments.
Observe that the tighter bound typically exceeds 70\% which is
significantly tighter than the a-priori bound of 1/2 for sequential planning.
Later, we will use these bounds to characterize solution quality across trials
with different planner configurations in lieu of comparison on common
subproblems.

\section{Experiment design}

\begin{figure}
  \begin{subfigure}[b]{0.24\linewidth}
    \includegraphics[width=\linewidth,trim={0 150 180 180},clip]{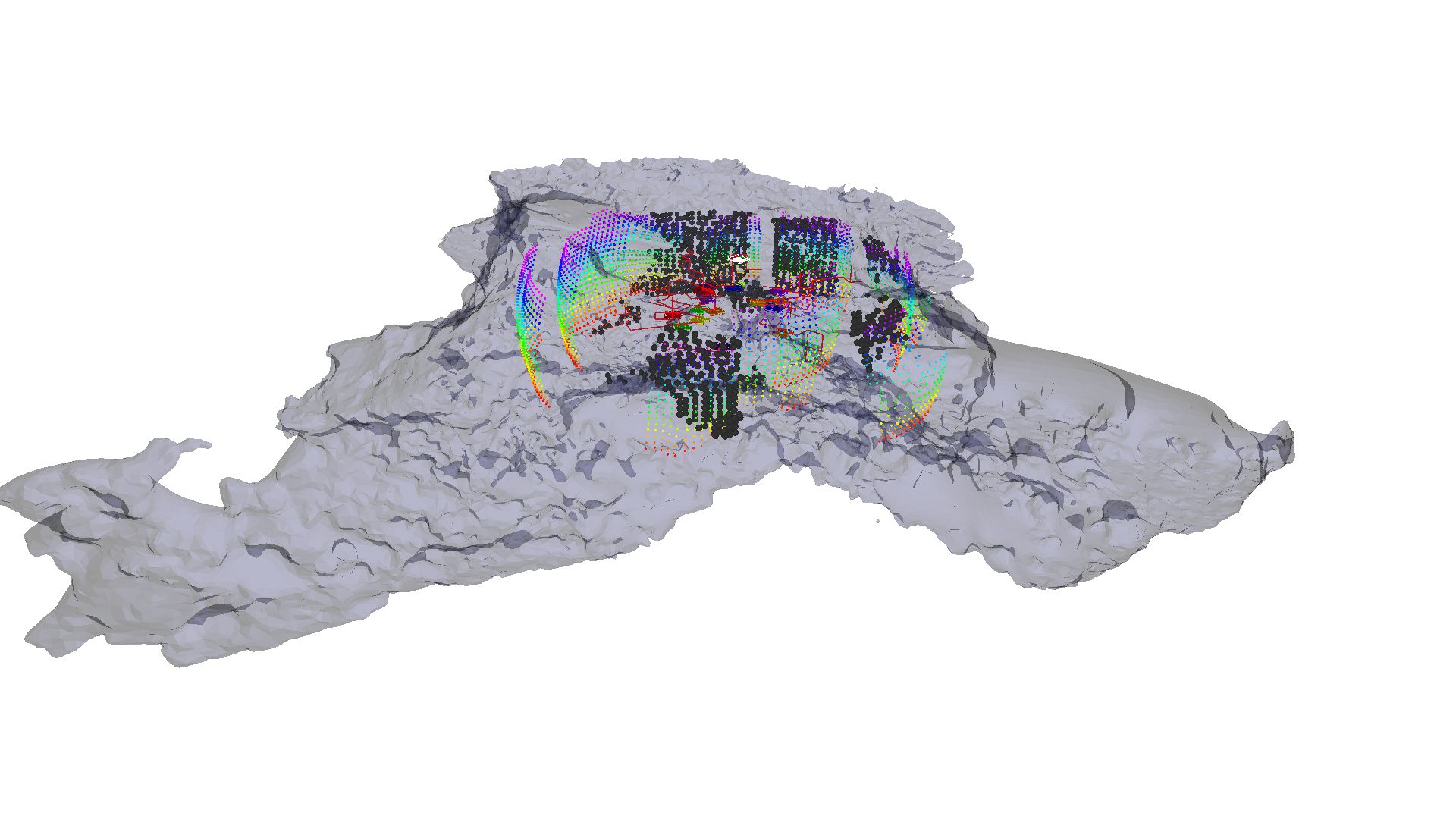}
    \caption{80 robot-iters.}
  \end{subfigure}
  \begin{subfigure}[b]{0.24\linewidth}
    \includegraphics[width=\linewidth,trim={0 150 180 180},clip]{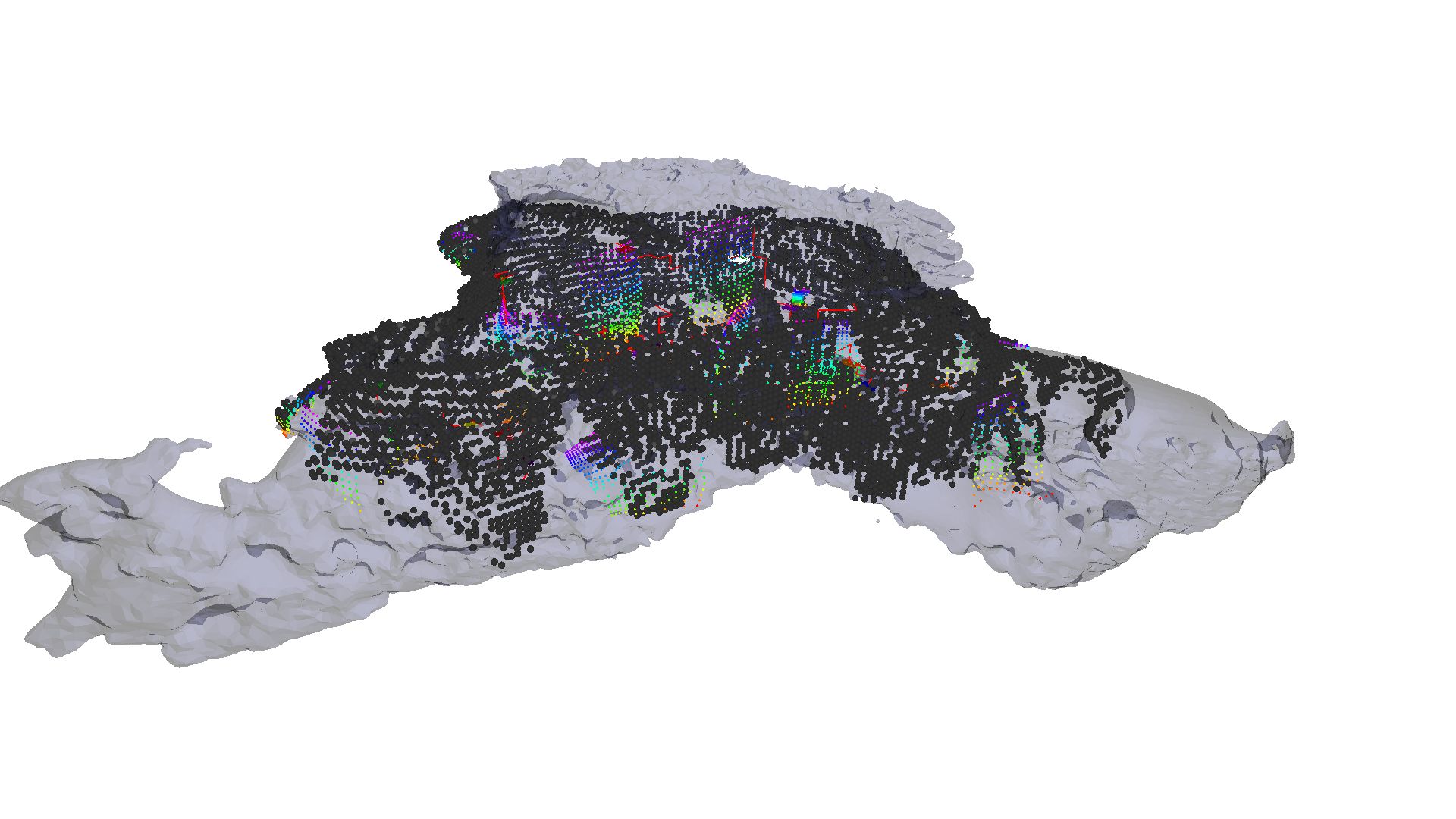}
    \caption{448 robot-iters.}
  \end{subfigure}
  \begin{subfigure}[b]{0.24\linewidth}
    \includegraphics[width=\linewidth,trim={0 150 180 180},clip]{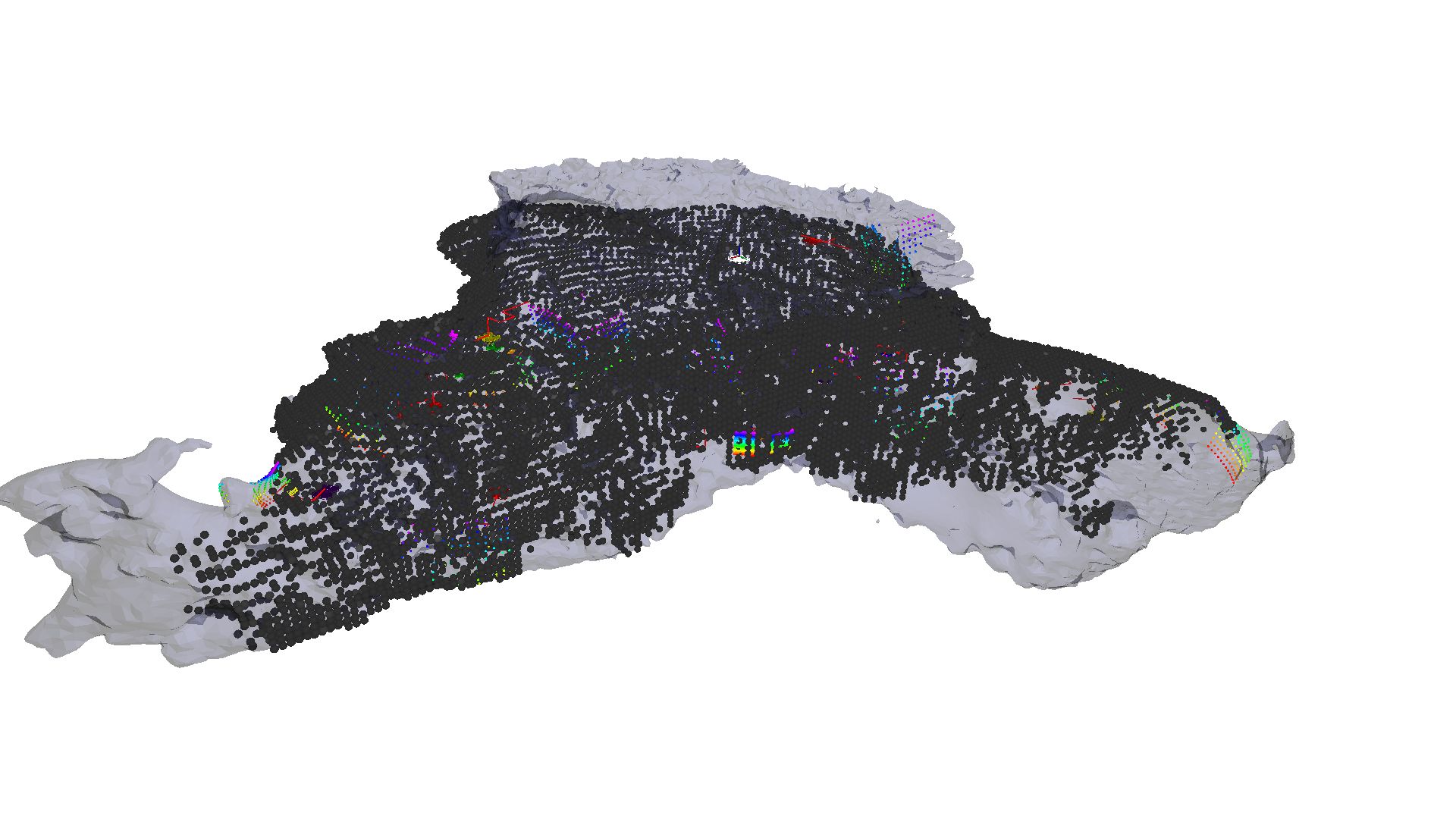}
    \caption{816 robot-iters.}
  \end{subfigure}
  \begin{subfigure}[b]{0.24\linewidth}
    \includegraphics[width=\linewidth,trim={0 150 180 180},clip]{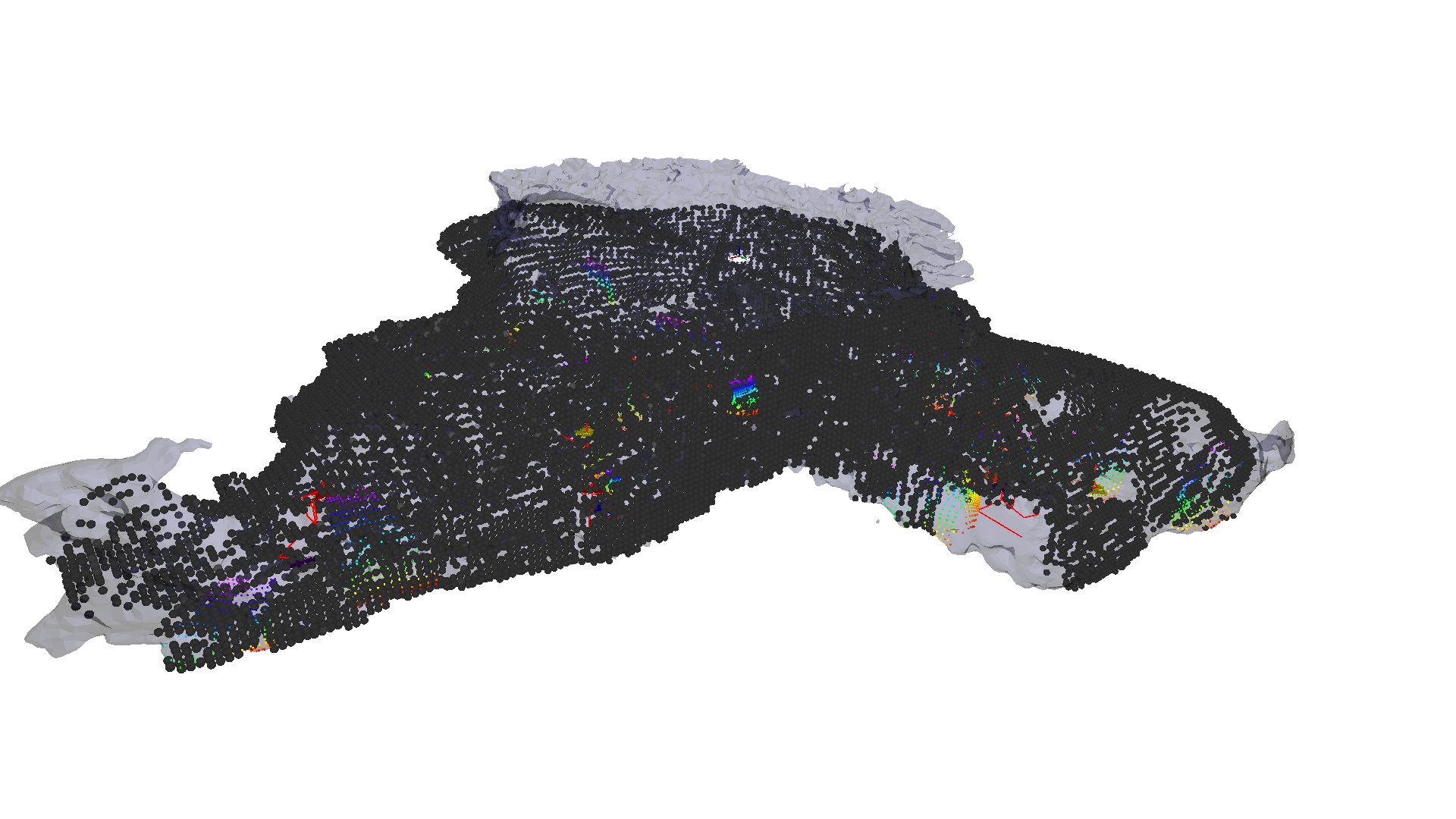}
    \caption{1184 robot-iters.}
  \end{subfigure}
  \caption[Visualization of exploration of the Skylight environment]{%
    The images above visualize the process of exploration of the
    Skylight environment with 16 robots and \rsp{} planning with $\numrounds=6$.
    Additionally, a video providing examples of the exploration process
    %for each environment and number of robots
    is available at:
    \url{https://youtu.be/B9j8LVIs384}
  }%
  \label{fig:time_sensitive_visualization}
\end{figure}

This section describes the design of the exploration experiments.
For intuition, Fig.~\ref{fig:time_sensitive_visualization} visualizes an example
of the exploration process and provides a link to a video providing examples for
all environments and numbers of robots.

\paragraph{Robot and sensor models}

The robot dynamics are based on a kinematic quadcopter model.
The set of control actions consists of $0.3\,\si{\metre}$ translations in the
cardinal directions with respect to the body frame as well as
$\pi/2\,\si\radian$ yawing motions.
Each robot obtains observations from a depth camera
with a range of $2.4\,\si\metre$, a resolution of
$19\times12$, and a field of view of $43.6\si\degree\times34.6\si\degree$.
Cameras face forward, oriented with the long axis vertical.

\paragraph{Single- and multi-robot planning}
\label{sec:planning_exploration}

Robots plan by collectively solving receding-horizon
planning problems \eqref{eq:receding_horizon_exploration}
with optimistic coverage rewards
(Sec.~\ref{sec:optimistic_coverage}).
They plan to maximize the objective individually
via Monte-Carlo tree search
(MCTS)~\citep{chaslot2010,browne2012,lauri2015ras,corah2019auro}
with 200 samples and collectively via specified methods for submodular
maximization
(Secs.~\ref{sec:sequential_planning} and~\ref{sec:rsp_planning}).
The planners also discount rewards, treating each robot as having an independent
probability of failure after each time-step.
Although we do not provide detail, the discounts
are compatible with the analysis of the objectives
(Secs.~\ref{sec:expected_coverage_exploration}
and~\ref{sec:mutual_information_exploration}), and evaluation remains
straightforward.

\paragraph{Environments and simulation scenarios}
\label{sec:environments_description}

The simulation results evaluate performance across a variety of environments
(listed in Table~\ref{tab:environments}).
In each case, robots start with random yaw and slightly perturbed positions near
a fixed starting location.
We determined maximum coverage values and the lengths of the
simulation trials (iterations per robot) through longer preliminary experiments
with conservative parameters.
%experiments with a low view value threshold ($\viewthreshold=100$)
%to encourage more complete exploration.
Additionally, all maps use a $0.1\,\si{\metre}$ discretization.
So, a volume of $1\,\si{\cubic\metre}$ contains 1000 grid cells.

\begin{table}
  \caption[List of exploration environments]{
    This table lists details of the test environments.
    While the \emph{bounding box volume} provides the volume of a bounding box
    around the environment, the \emph{exploration volume} lists the approximate
    maximum environment coverage volume for exploration.
    %Images portray partially explored environments with unknown space
    %(excluding subterranean environments)
    %in gray
    %and occupied in black.
  }%
  \label{tab:environments}
  \centering
  \setlength{\figurewidth}{0.18\linewidth}
  % Somehow the below shifts things to the midline
  \renewcommand{\tabularxcolumn}[1]{>{\arraybackslash}m{#1}}
  \footnotesize
  \begin{tabularx}{\linewidth}{lXcc}
    Image & Name
          & \pbox{14ex}{Bounding Box \\ Volume}
          & \pbox{12ex}{Exploration Volume}
    \\\belowtoprule
    %\\
    %%
    %% Title comment
    %%
    %\raisebox{-0.5\height}{%
    % \includegraphics[width=\figurewidth]{environments/#}%
    %} % Image
    %& Name
    %& $0\,\si{\cubic\metre}$ % Bounding box volume
    %& $0\,\si{\cubic\metre}$ % Exploration volume
    %& \scriptsize
    % Description
    %%%%%%%%%%%%%%%%%%%%%%%%%%%%%%%%%%%%%%%%%%%%%%%%%%%%%%%%%%%%%%%%%
    %
    % Boxes
    %
    \raisebox{-0.5\height}{%
      \includegraphics[width=0.8\figurewidth]{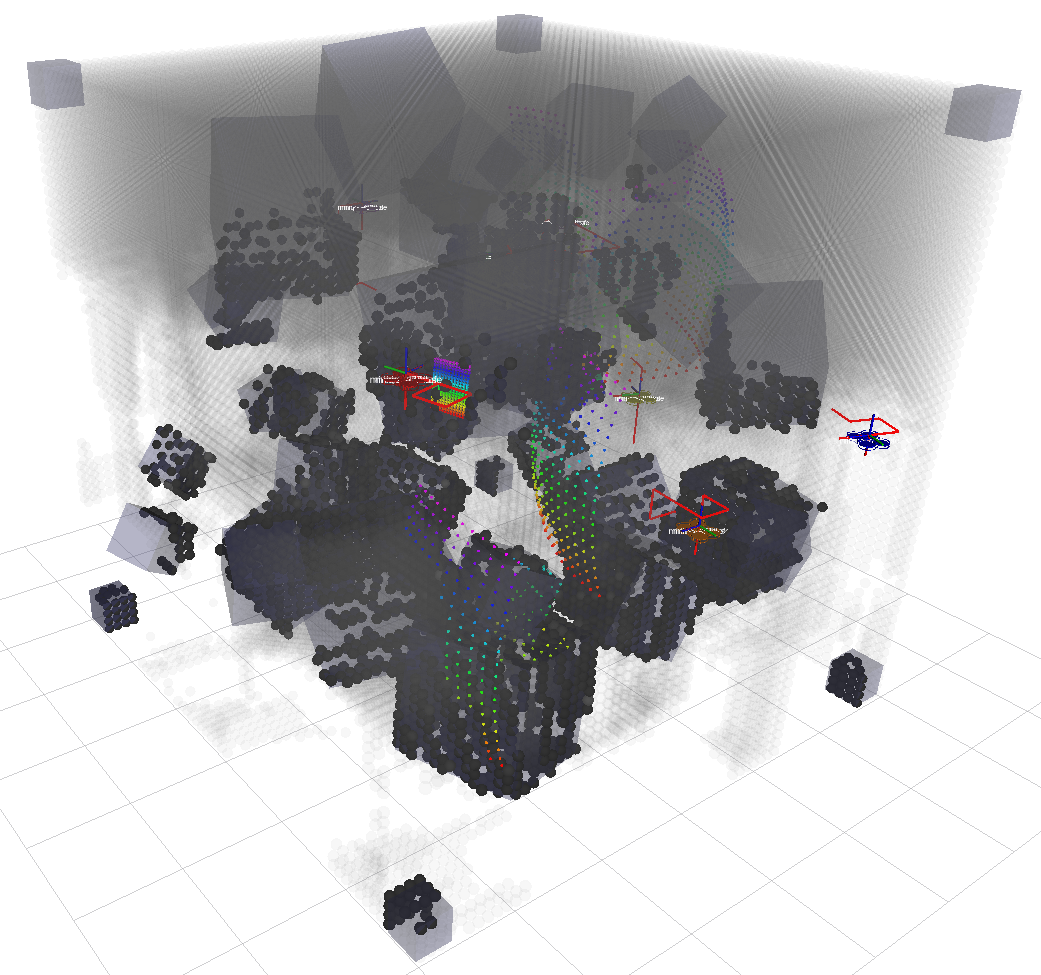}%
    } % Image
     & Boxes
     & $216\,\si{\cubic\metre}$ % Bounding box volume
     & $199\,\si{\cubic\metre}$ % Exploration volume
     %& \scriptsize
     %Scattered boxes cause occlusions in a $6\,\si\metre$ cube.
     %Robots start offset at bottom and move \emph{upward}.
    \\
    %%%%%%%%%%%%%%%%%%%%%%%%%%%%%%%%%%%%%%%%%%%%%%%%%%%%%%%%%%%%%%%%%
    %
    % Hallway-Boxes
    %
    \\
    \raisebox{-0.5\height}{%
     \includegraphics[width=\figurewidth]{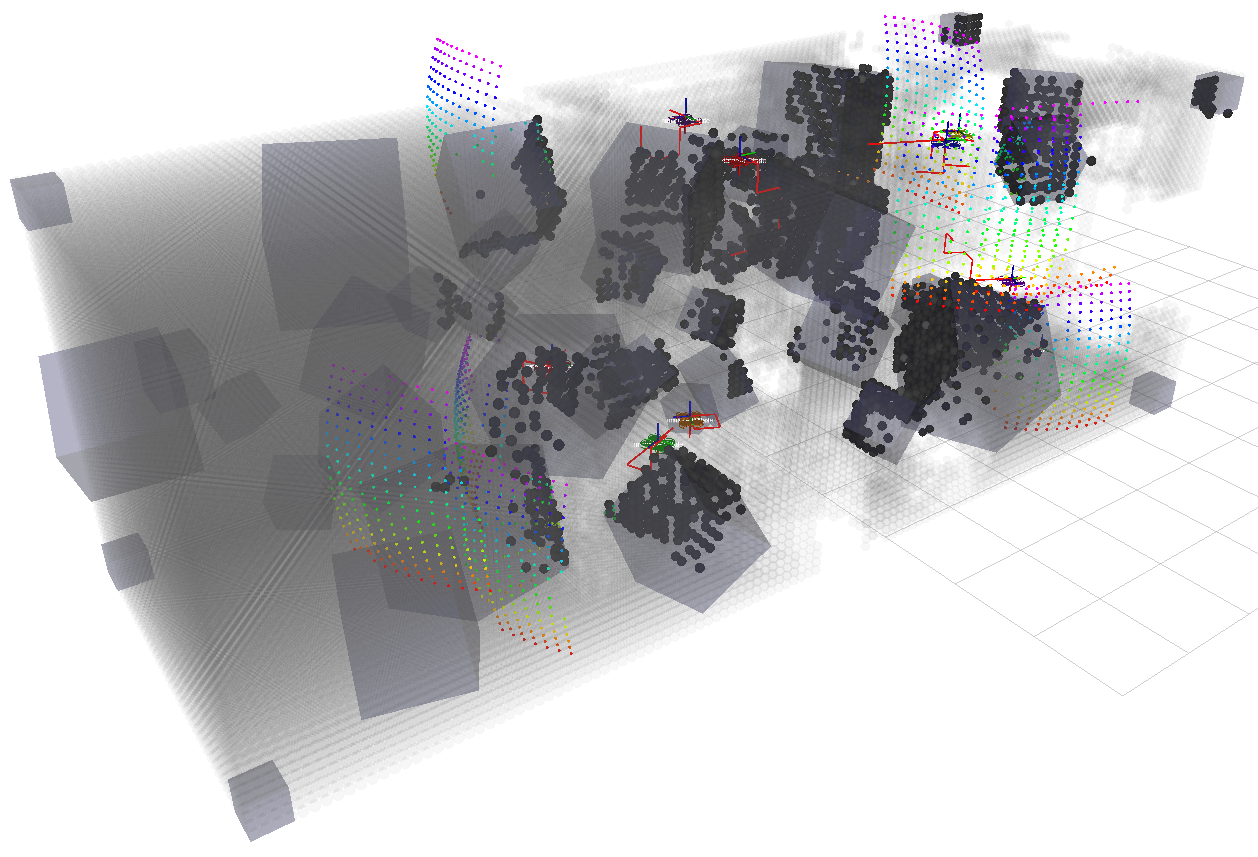}%
    } % Image
    & Hallway-Boxes
    & $217\,\si{\cubic\metre}$ % Bounding box volume
    & $202\,\si{\cubic\metre}$ % Exploration volume
    %& \scriptsize
    %A rearrangement of the boxes environment into a $12\,\si{\metre}$ square
    %prism.
    %Robots start at one end and move toward the other.
    \\
    %%%%%%%%%%%%%%%%%%%%%%%%%%%%%%%%%%%%%%%%%%%%%%%%%%%%%%%%%%%%%%%%%
    %
    % Plane-Boxes
    %
    \raisebox{-0.5\height}{%
     \includegraphics[width=\figurewidth]{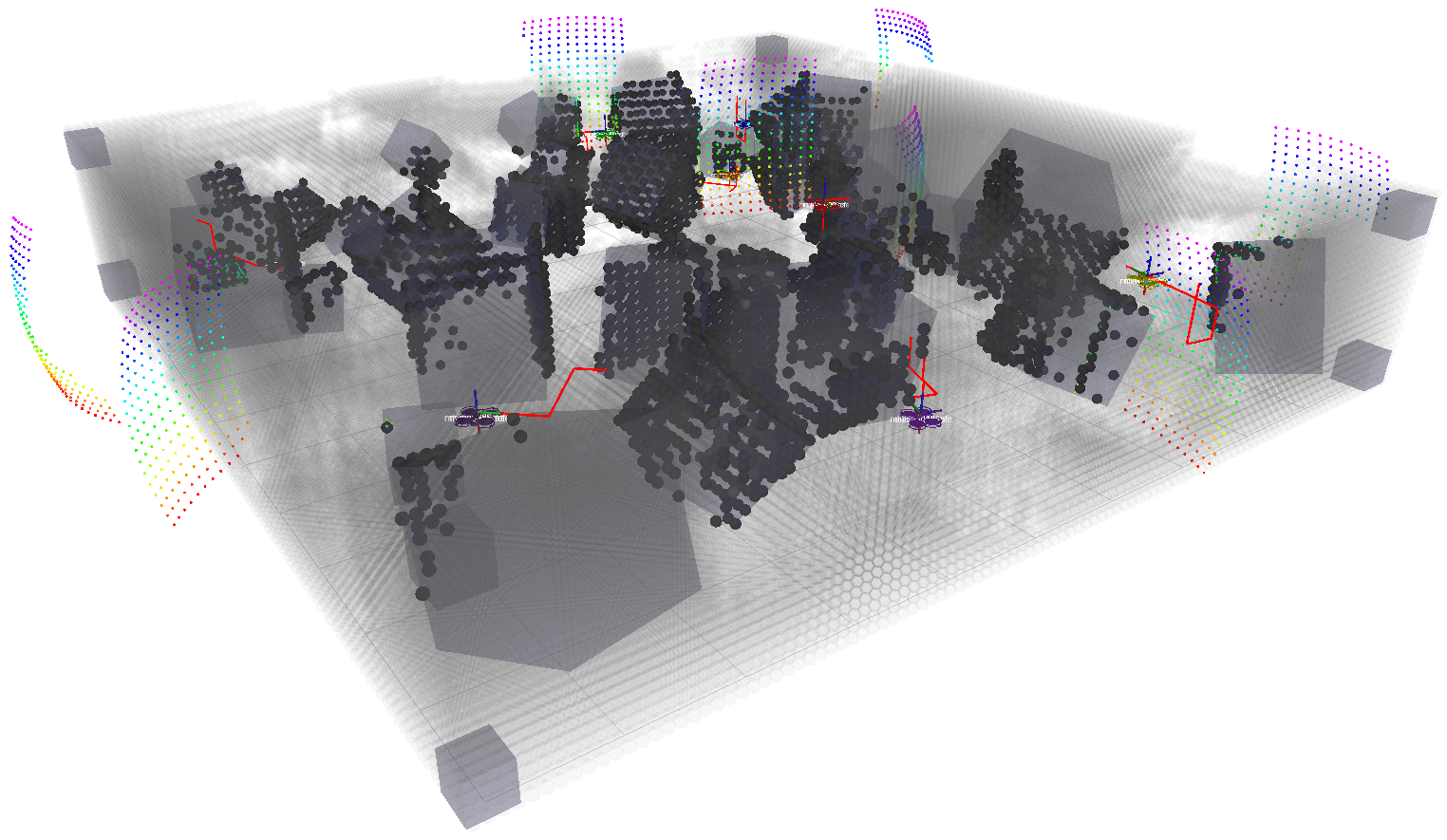}%
    } % Image
    & Plane-Boxes
    & $227\,\si{\cubic\metre}$ % Bounding box volume
    & $212\,\si{\cubic\metre}$ % Exploration volume
    %& \scriptsize
    %Rearrangement of the boxes environment into $2\,\si{\metre}$ tall square
    %planar configuration, and robots start in the center.
    %This highlights performance in common, primarily two-dimensional
    %environments.
    \\
    %%%%%%%%%%%%%%%%%%%%%%%%%%%%%%%%%%%%%%%%%%%%%%%%%%%%%%%%%%%%%%%%%
    %
    % Empty
    %
    \raisebox{-0.5\height}{%
     \includegraphics[width=\figurewidth]{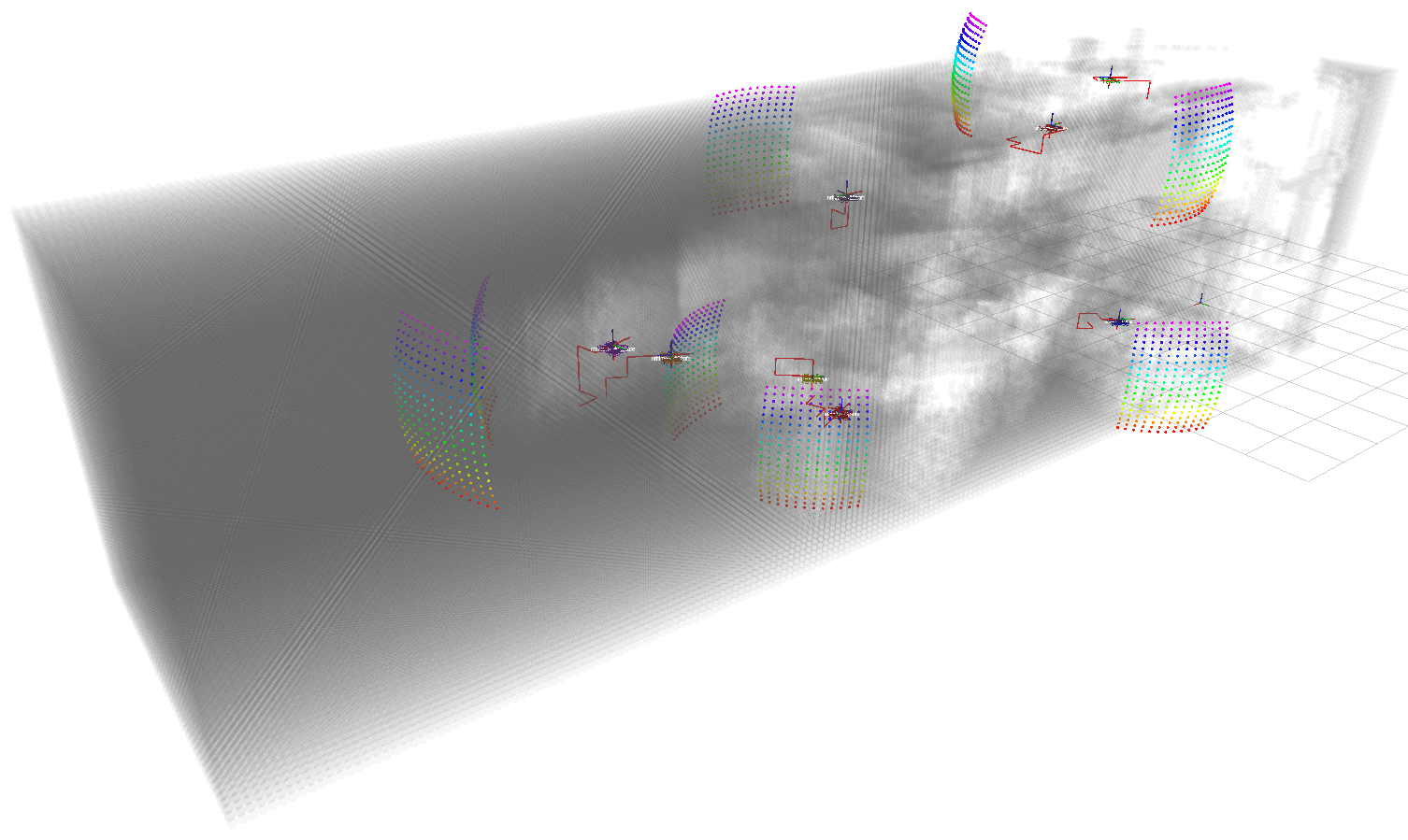}%
    } % Image
    & Empty
    & $500\,\si{\cubic\metre}$ % Bounding box volume
    & $500\,\si{\cubic\metre}$ % Exploration volume
    %& \scriptsize
    %Robots start on one end of a $20\,\si{\metre}$ hallway-like square prism
    %that is completely devoid of obstacles which highlight steady-state
    %performance in open space.
    \\
    %%%%%%%%%%%%%%%%%%%%%%%%%%%%%%%%%%%%%%%%%%%%%%%%%%%%%%%%%%%%%%%%%
    %
    % Skylight
    %
    \raisebox{-0.5\height}{%
     \includegraphics[width=\figurewidth]{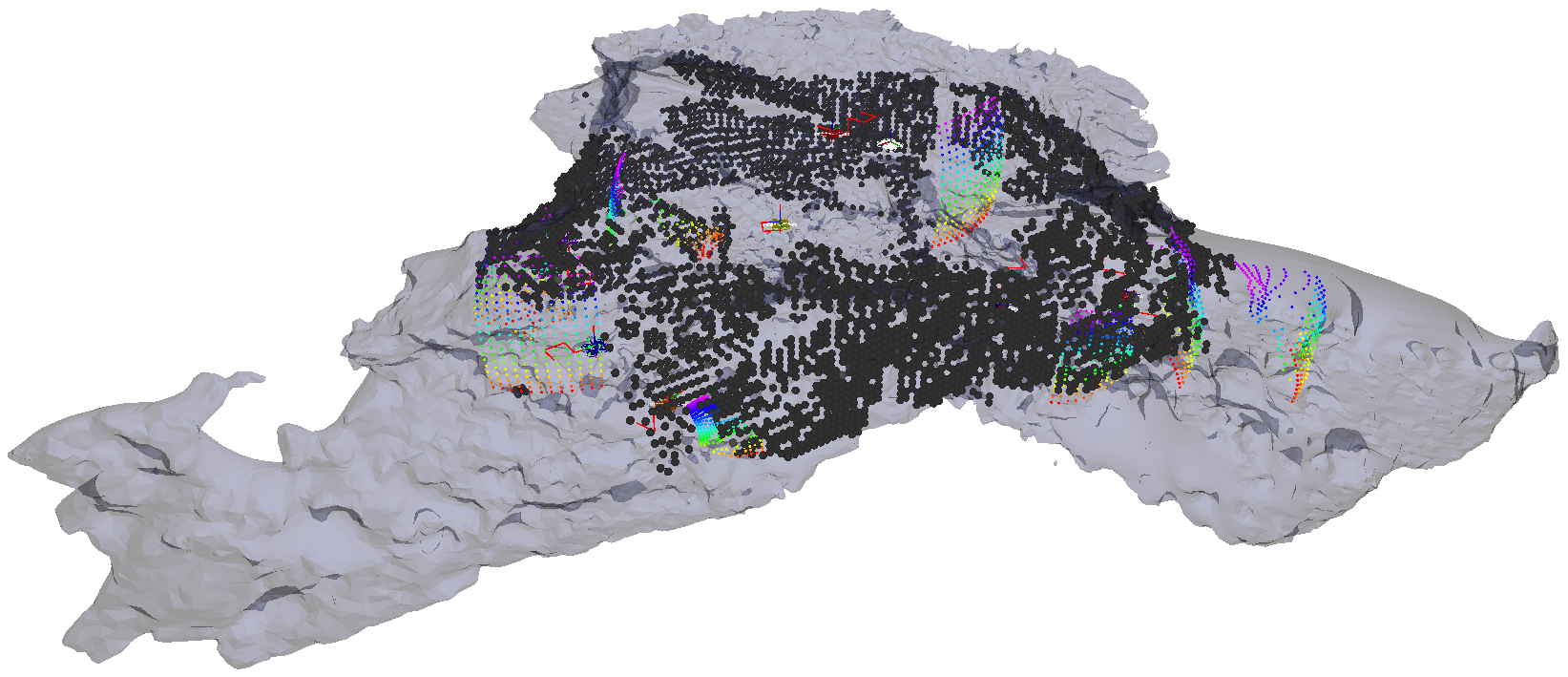}%
    } % Image
    & Skylight
    & N/A % Bounding box volume
    & $220\,\si{\cubic\metre}$ % Exploration volume
    %& \scriptsize
    %Mesh based on survey data%
    %~\citep{pits_and_caves,jones2015,jones2016phd,tabib2016iros}
    %from the Indian Tunnel
    %skylight at Craters of the Moon National Park, scaled down to $\sim\!35\%$
    %of actual size.
    %Robots start at the top of the mouth (now $7\,\si{\metre}$ in diameter).
    \\
    %%%%%%%%%%%%%%%%%%%%%%%%%%%%%%%%%%%%%%%%%%%%%%%%%%%%%%%%%%%%%%%%%
    %
    % Office (server room)
    %
    \raisebox{-0.5\height}{%
     \includegraphics[width=0.9\figurewidth]{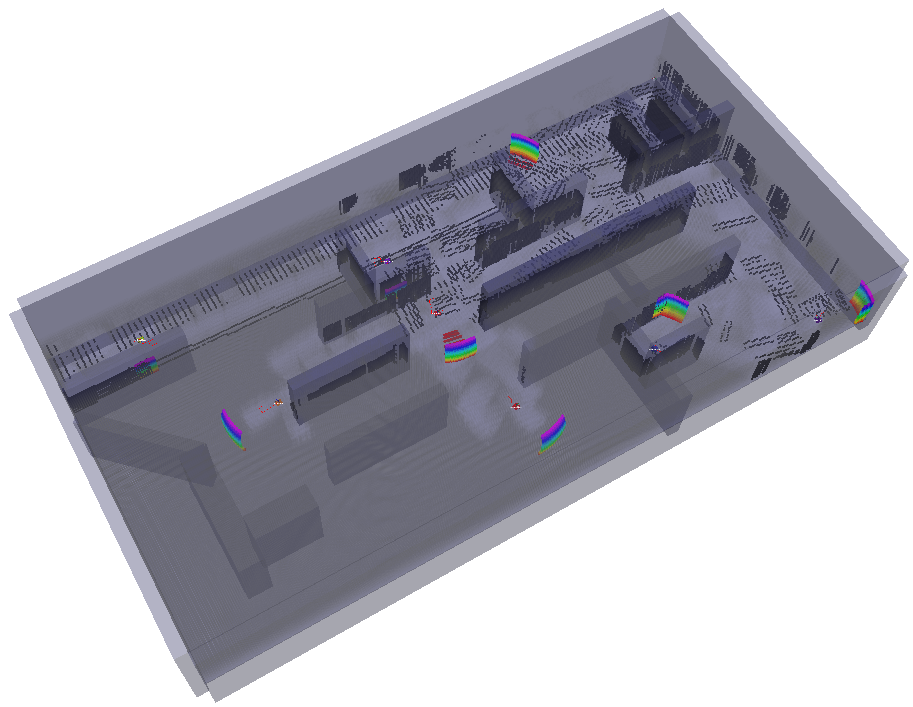}%
    } % Image
    & Office
    & $1300\,\si{\cubic\metre}$ % Bounding box volume (0mx0mx2m
    & $1180\,\si{\cubic\metre}$ % Exploration volume
    %& \scriptsize
    %A simulated
    %$36\,\si{\metre}\times
    %18\,\si{\metre}\times
    %2\,\si{\metre}$
    %office environment.
    %This environment almost maze-like in design, and the height
    %effectively restricts robots to planar motion.
    %Robots start in the upper right corner.
%   \\
%   %%%%%%%%%%%%%%%%%%%%%%%%%%%%%%%%%%%%%%%%%%%%%%%%%%%%%%%%%%%%%%%%
%   %
%   % Rapps cave
%   %
%   \raisebox{-0.5\height}{%
%    \includegraphics[width=\figurewidth]{environments/rapp_cave}%
%   } % Image
%   & Rapps-Cave
%   & N/A
%   & $1060\,\si{\cubic\metre}$ % Exploration volume
%   & \scriptsize
%   Reconstruction of the Rapps cave~\citep{tabib2019fsr}. The environment is
%   scaled down by 60\%, now $40\,\si{\metre}$ in the largest dimension.
%   Robots start on the end toward left.
    \\\bottomrule
  \end{tabularx}
\end{table}

\section{Results: Varying the distributed planner}

\begin{table}
  \caption[Planner parameters for receding-horizon exploration]{%
    Planner parameters for receding-horizon exploration.
    The myopic and sequential planners were tuned separately to maximize
    performance for 16 robots in the Boxes and Empty environments
    (Table.~\ref{tab:environments}).
    The parameter $c_p$ belongs to the MCTS planner~\citep{browne2012}
    and is set to roughly half the typical value of the objective for a
    single robot.
    \emph{All \rsp{} planners use parameters for sequential planning.}
  }
  \renewcommand{\tabularxcolumn}[1]{>{\arraybackslash}m{#1}}
  \scriptsize
  \label{tab:exploration_parameters}
  \begin{tabularx}{\linewidth}{Xlllll}
    Planner
    & $c_p$
    & \pbox{9ex}{Horizon ($L$)}
    & \pbox{11ex}{View Value \\ Threshold ($\viewthreshold$)}
    & \pbox{11ex}{View Distance Factor ($\distancefactor$)}
    & \pbox{9ex}{Discount Factor}
    \\\belowtoprule
    Sequential
               & 1500 % c_p
               & 10 % Horizon
               & 900 % View threshold
               & 500 % Distance factor
               & 0.7 % Discount factor
    \\
    Myopic
               & 1500 % c_p
               & 10 % Horizon
               & 300 % View threshold
               & 700 % Distance factor
               & 1.0 % Discount factor
    \\\bottomrule
  \end{tabularx}
\end{table}

The first study evaluates the effect of the method of multi-robot coordination
on exploration performance.
The results compare sequential planning (Sec.~\ref{sec:sequential_planning}),
myopic planning
(wherein robots plan via MCTS and ignore others' decisions),
and \rsp{} with
1, 3, and 6 rounds ($\numrounds$).\footnote{%
  Except, we provide results for only $\numrounds=6$
  for the larger Office environment due to longer trial time.
}
Parameters
(see Table~\ref{tab:exploration_parameters})
were selected separately for myopic and sequential planners with
\rsp{} inheriting parameters for sequential planning.\footnote{
  \rsp{} with $\numrounds=1$ is equivalent to myopic planning but will
  use the same parameters as sequential planning so that any adverse impacts of
  parameter selection on the myopic planner will be evident.
}
We provide results for 10 trials per each configuration, varying planners,
environments, and numbers of robots (4, 8, 16, and 32).

\begin{figure}
  \tiny
  \begin{subfigure}[b]{0.49\linewidth}
    \setlength{\figurewidth}{1.24\linewidth}
    \setlength{\figureheight}{0.9\linewidth}
    \input{./fig/environment_coverage/thesis_experiments_coverage_boxes_2420f96.tex}
    \vspace{-0.4cm}
    \caption{Boxes}
  \end{subfigure}
  \begin{subfigure}[b]{0.49\linewidth}
    \setlength{\figurewidth}{1.24\linewidth}
    \setlength{\figureheight}{0.9\linewidth}
    \input{./fig/environment_coverage/thesis_experiments_coverage_hallway_2420f96.tex}
    \vspace{-0.2cm}
    \caption{Hallway-Boxes}
  \end{subfigure}
  \begin{subfigure}[b]{0.49\linewidth}
    \setlength{\figurewidth}{1.24\linewidth}
    \setlength{\figureheight}{0.9\linewidth}
    \input{./fig/environment_coverage/thesis_experiments_coverage_plane_2420f96.tex}
    \vspace{-0.4cm}
    \caption{Plane-Boxes}
  \end{subfigure}
  \hspace{0.05cm}
  \begin{subfigure}[b]{0.49\linewidth}
    \setlength{\figurewidth}{1.24\linewidth}
    \setlength{\figureheight}{0.9\linewidth}
    \hspace{0.1cm}
    \input{./fig/environment_coverage/thesis_experiments_coverage_empty_2420f96.tex}
    \vspace{-0.4cm}
    \caption{Empty}
  \end{subfigure}
  \begin{subfigure}[b]{0.49\linewidth}
    \setlength{\figurewidth}{1.24\linewidth}
    \setlength{\figureheight}{0.9\linewidth}
    \input{./fig/environment_coverage/thesis_experiments_coverage_indian_tunnel_skylight_2420f96.tex}
    \vspace{-0.3cm}
    \caption{Skylight}
  \end{subfigure}
  \begin{subfigure}[b]{0.49\linewidth}
    \setlength{\figurewidth}{1.24\linewidth}
    \setlength{\figureheight}{0.9\linewidth}
    \input{./fig/environment_coverage/thesis_experiments_coverage_server_room_scaled_0_6_2420f96.tex}
    \vspace{-0.4cm}
    \caption{Office}
  \end{subfigure}
  \caption[Environment coverage results]{Environment coverage results for each
    environment with different planners.
    Black lines demarcate the maximum environment coverage for a given
    environment and dashed black lines mark a threshold at 90\% of that value to
    represent task completion.
    Note, total coverage rates always increase with numbers of robots
    despite decrease in efficiency evident here.
  }%
  \label{fig:environment_coverage}
\end{figure}
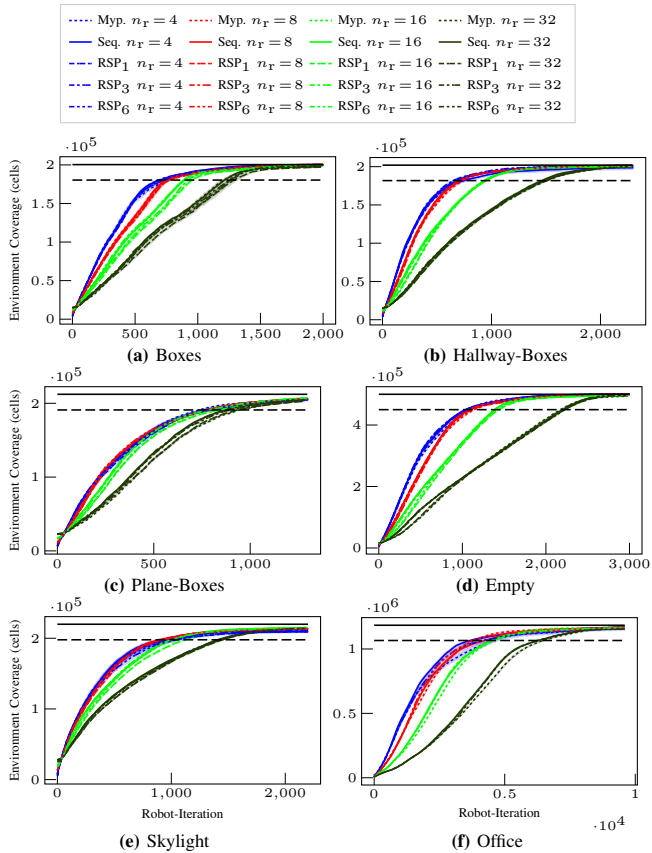

Figure~\ref{fig:environment_coverage} summarizes the exploration process
for these simulations
in terms of environment coverage
\eqref{eq:environment_coverage}.
%and highlights the maximum coverage values (Table~\ref{tab:environments}).
Although, there is not always much variation across planner configurations,
these results also illustrate consistency in coverage rates, graceful
degradation in efficiency (per-robot) with increasing numbers, and reliably
complete exploration.
The latter, reliable completion, is evident in convergence
toward the maximum environment coverage
and low variance toward the end of exploration trials.

\subsection{Online bounds on suboptimality}

\begin{figure}
  \tiny
  \begin{subfigure}[b]{0.32\linewidth}
    \setlength{\figurewidth}{1.38\linewidth}
    \setlength{\figureheight}{1.0\figurewidth}
    \input{./fig/suboptimality/thesis_experiments_suboptimality_boxes_2420f96.tex}
    \vspace{-0.3cm}
    \caption{Boxes}
  \end{subfigure}
  \hspace{0.1cm}
  \begin{subfigure}[b]{0.32\linewidth}
    \setlength{\figurewidth}{1.38\linewidth}
    \setlength{\figureheight}{1.0\figurewidth}
    \input{./fig/suboptimality/thesis_experiments_suboptimality_hallway_2420f96.tex}
    \vspace{-0.3cm}
    \caption{Hallway-Boxes}
  \end{subfigure}
  \hspace{-0.1cm}
  \begin{subfigure}[b]{0.32\linewidth}
    \setlength{\figurewidth}{1.38\linewidth}
    \setlength{\figureheight}{1.0\figurewidth}
    \input{./fig/suboptimality/thesis_experiments_suboptimality_plane_2420f96.tex}
    \vspace{-0.3cm}
    \caption{Plane-Boxes}
  \end{subfigure}
  \begin{subfigure}[b]{0.32\linewidth}
    \setlength{\figurewidth}{1.38\linewidth}
    \setlength{\figureheight}{1.0\figurewidth}
    \input{./fig/suboptimality/thesis_experiments_suboptimality_empty_2420f96.tex}
    \vspace{-0.3cm}
    \caption{Empty}
  \end{subfigure}
  \hspace{0.1cm}
  \begin{subfigure}[b]{0.32\linewidth}
    \setlength{\figurewidth}{1.38\linewidth}
    \setlength{\figureheight}{1.0\figurewidth}
    \input{./fig/suboptimality/thesis_experiments_suboptimality_indian_tunnel_skylight_2420f96.tex}
    \vspace{-0.3cm}
    \caption{Skylight}
  \end{subfigure}
  \hspace{-0.1cm}
  \begin{subfigure}[b]{0.32\linewidth}
    \setlength{\figurewidth}{1.38\linewidth}
    \setlength{\figureheight}{1.0\figurewidth}
    \input{./fig/suboptimality/thesis_experiments_suboptimality_server_room_scaled_0_6_2420f96.tex}
    \vspace{-0.3cm}
    \caption{Office}
  \end{subfigure}
  \caption[Lower bounds on suboptimality for receding-horizon planning for
  exploration]{%
    Lower bounds on suboptimality for receding-horizon planning for exploration
    in each environment.
    Plots provide mean values and standard error
    (\emph{w.r.t.} the standard deviation of trial means)
    for data up to completion of the exploration task.
    These results demonstrate how the performance of \rsp{} planning approaches
    that of sequential planning with increasing numbers of planning rounds from
    $\numrounds=1$
    to
    $\numrounds=6$.
    We exclude the Myopic planner as the suboptimality bounds are not
    necessarily comparable to those of the other planners because the choice of
    parameters (Table~\ref{tab:exploration_parameters})
    changes the objective values.
    Instead,
    \rsp[1]{} also plans myopically but has the same problem parameters as other
    planner configurations.
  }%
  \label{fig:exploration_suboptimality}
\end{figure}
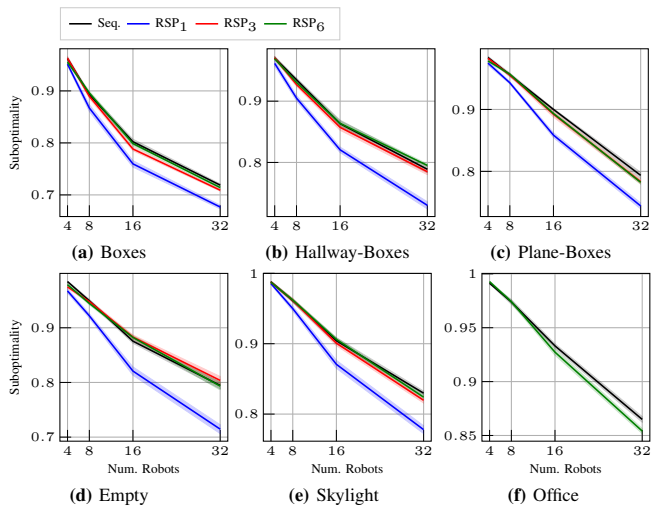

Figure~\ref{fig:exploration_suboptimality} presents results on mean values of
the lower bounds\footnote{
  Bounds computed approximately via MCTS (single-robot planning)
  and are only representative of suboptimality in multi-robot coordination.
}
on suboptimality
(the greater of the two bounds from Sec.~\ref{sec:online_bounds_exploration}).
These plots demonstrate that the suboptimality of \rsp{} planning
approaches sequential planning (Sec.~\ref{sec:sequential_planning}) with
increasing numbers of planning rounds ($\numrounds$)
as the performance bounds for these planners suggest\footnote{
  Strictly, the bounds for \rsp{} planning only establish convergence to the
  same worst case suboptimality as sequential planning (1/2), but we expect
  comparable suboptimality in practice.
}~\citep{corah2018cdc}.
Additionally, the actual suboptimality is consistently better than the worst
case bound of 1/2 for sequential planning
which is consistent with observations from related
works~\citep{leskovec2007,golovin2011jair,krause2008}.
Still, solutions for larger numbers of robots are more suboptimal for all
planners and exhibit greater differences in suboptimality
(reaching 8\% for Empty).
The decrease in these bounds with increasing numbers of robots is representative
of robots operating in closer proximity and with greater overlap between
observations over the planning horizon.
As a whole, these results demonstrate \rsp{} providing solution quality
comparable to sequential planning with only 3 or 6 computation steps versus as
many as 32 for sequential planning.
%This decrease is also not necessarily representative of commensurate changes in
%actual suboptimality (as we present statistics for \emph{lower bounds})
%and may simply reflect effects of changing conditions.

\section{Results: Varying the objective}

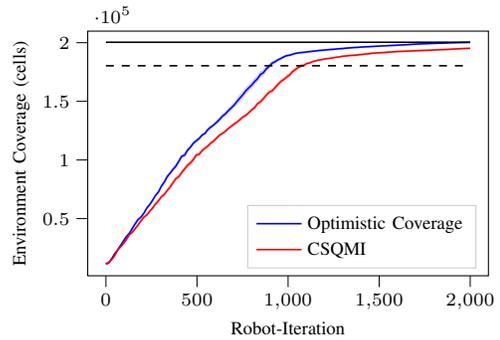
\begin{figure}
  \centering
  \scriptsize
  \setlength{\figurewidth}{0.8\linewidth}
  \setlength{\figureheight}{0.7\figurewidth}
  \input{fig/test_csqmi_coverage_coverage_boxes_afdb01c.tex}
  \caption{%
    The above compares exploration with optimistic coverage and
    information-based CSQMI
    objectives~\citep{charrow2015icra} in terms of environment coverage in the
    Boxes environment.
    %(with 16 robots and ten trials of sequential planning).
    Black lines demarcate the maximum environment coverage,
    and a threshold at 90\% to represent task completion.
  }%
  \label{fig:csqmi_and_coverage}
\end{figure}

Now, let us consider the effect of the choice of objective.
This study, compares optimistic coverage
(Sec.~\ref{sec:optimistic_coverage}) to an information-based objective,
Cauchy-Schwarz quadratic mutual information (CSQMI)~\citep{charrow2015icra} with
an occupancy prior of 0.125.
Recall that optimistic coverage is equivalent to mutual information in the limit
for small priors and is accurate for multiple rays and camera views.
On the other hand, CSQMI and other information-based
objectives~\citep{charrow2015icra,julian2014,zhang2019ijrr,henderson2020icra}
can evaluate individual rays accurately but rely on approximations for
collections of rays and views.
Figure~\ref{fig:csqmi_and_coverage} plots the environment coverage for ten
trials with sixteen robots in the Boxes environment.
We compare performance in terms of time to complete the exploration task,
defined as the time to reach 90\% of the maximum exploration volume
(see Tab.~\ref{tab:exploration_parameters}).

Interestingly, robots planning with optimistic coverage explore the environment
16\% faster than with CSQMI (averaging 890 robot-iterations versus 1067
respectively).
While this result is limited in scope, the significant difference in
performance suggests that accurate evaluation for multiple views may be more
important than accurate evaluation of information gain for individual rays for
robotic exploration.

\section{Conclusion and future work}

This work has studied multi-robot exploration of three-dimensional environments
from the perspective of design of objectives that quantify collections of camera
views to reward observation of unknown space.
Establishing that mutual information without noise is a special case of expected
coverage enabled us to re-interpret coverage objectives as limiting cases of
mutual information.
Toward this end, simulation results found that employing a coverage-based
objective improved completion time by 16\% compared to a ray-based
approximation of mutual information~\citep{charrow2015icra}.
In future work, these results and our analysis may also produce improvements
in approximation of mutual information based on expected coverage
(see discussion in~\citep{corah2020phd}).

The analysis also overcame an important challenge for distributed,
receding-horizon planning for multi-robot exploration.
By proving that mutual information without noise is expected coverage,
we proved that this case satisfies suboptimality guarantees for distributed
planning via Randomized Sequential Partitions (\rsp{})~\citep{corah2018cdc}.
\rsp{} runs with a fixed numbers of sequential steps rather than one per
robot, unlike existing sequential methods~\citep{singh2009,fisher1978}, so this
is a significant result for distributed exploration.
Additionally, the results demonstrated that distributed submodular maximization
can provide consistent improvements in suboptimality on receding-horizon
subproblems.

% Fakesection: Bibliography
{
\bibliographystyle{IEEEtranN}
%\small
\bibliography{IEEEabrv,bibliography}
}

\newpage

\ifextended
\appendices

\section{Proof of Theorem~\ref{theorem:mapping_information_is_coverage}}
\label{appendix:proof_that_mapping_information_is_coverage}

The following proves that noiseless mutual information with independent cells is
3-increasing by taking advantage of cell independence liberally to write mutual
information in terms of the expected entropy of the cells that the robots will
observe.

\begin{proof}
  We can write the mutual information\footnote{
    For more detail on information theory, please refer to \citet{cover2012}.
  }
  between
  the environment $\environment$
  and
  future observations $\observations(X)$
  in terms of entropies:
  % Expected mutual information for covered cells
  \begin{align}
    \MI(\environment; \observations(X))
    &=
    \H(\environment) - \H(\environment | \observations(X)).
  \end{align}
  The conditional entropy can then be rewritten
  in terms of the expected entropy
  %\eqref{eq:conditional_entropy_expectation}
  given the direct observations of
  cell occupancy \eqref{eq:exploration_observation}
  associated with a hypothetical instantiation of the environment
  $\environment'$
  while
  abbreviating observed cells as $C'=\covercells(X,\environment')$:
  \begin{align}
    \MI(\environment; \observations(X))
    &=
    \H(\environment)
    -
    \E_{\environment' \sim \environmentsguess}
    \left[
      \H(\environment |
      \environment_{C'}\! =\! \environment'_{C'}
      )
    \right].
  \end{align}
  Then, observing that conditional entropy is simply the entropy of the cells
  that have not yet been observed
  $D' = \cells \setminus C'$,
  due to independence:
  \begin{align}
    \MI(\environment; \observations(X))
    &=
    \H(\environment)
    -
    \E_{\environment' \sim \environmentsguess}
    \left[
      \H(\environment_{D'})
    \right].
  \end{align}
  Next, bringing the entropy of $\environment$ into the expectation
  does not change its value,
  and separating the independent observed and unobserved cells
  simplifies the expression:
  \begin{align}
    \MI(\environment; \observations(X))
    &=
    \E_{\environment' \sim \environmentsguess}
    \left[
      \H(\environment_{C'})
      +
      \H(\environment_{D'})
      -
      \H(\environment_{D'})
    \right].
    \\
    &=
    \E_{\environment' \sim \environmentsguess}
    \left[
      \H(\environment_{C'})
    \right].
\end{align}

Finally, the joint entropy of the cells the robot will observe
is the sum of their individual entropies
\begin{align}
  \MI(\environment; \observations(X)) &=
  \E_{\environment' \sim \environmentsguess}
  \left[
    \sum\nolimits_{i \in \covercells(X, \environment')} \H(\cell_i)
  \right].
\end{align}
This expresses a weighted expected coverage objective
\eqref{eq:expected_coverage} where the weight $\cellweight(i)$ of each cell $i
\in \cells$ is equal to its entropy $\H(\cell_i)$.
Observing that weighted expected coverage is 3-increasing
(Sec.~\ref{sec:expected_coverage_monotonicity})
completes the proof.
\end{proof}
\else\fi

\end{document}

%% file: fig/environment_coverage/thesis_experiments_coverage_boxes_2420f96.tex
% This file was created by matplotlib2tikz v0.7.6.
\begin{tikzpicture}

\definecolor{color0}{rgb}{0.12156862745098,0.227450980392157,0.00392156862745098}

\begin{axis}[
height=\figureheight,
legend cell align={left},
legend columns=4,
legend style={at={(0.00,1.20)}, anchor=south west, draw=white!80.0!black},
legend image post style={scale=0.5}, % resize legend entries
tick align=inside,
tick pos=left,
width=\figurewidth,
x grid style={lightgray!92.0261437908!black},
xmin=-100, xmax=2100,
xtick style={color=black},
y grid style={lightgray!92.0261437908!black},
ylabel={Environment Coverage (cells)},
ymin=-6383.79138812876, ymax=210614.877920107,
ytick style={color=black}
]
% Axis content deleted
% See actual content in: <filename>.tex.bac
\end{axis}

\end{tikzpicture}

%% file: fig/environment_coverage/thesis_experiments_coverage_hallway_2420f96.tex
% This file was created by matplotlib2tikz v0.7.6.
\begin{tikzpicture}

\definecolor{color0}{rgb}{0.12156862745098,0.227450980392157,0.00392156862745098}

\begin{axis}[
height=\figureheight,
tick align=inside,
tick pos=left,
width=\figurewidth,
x grid style={lightgray!92.0261437908!black},
xmin=-115, xmax=2415,
xtick style={color=black},
y grid style={lightgray!92.0261437908!black},
ymin=-6779.41482546816, ymax=211905.210229784,
ytick style={color=black}
]
% Axis content deleted
% See actual content in: <filename>.tex.bac
\end{axis}

\end{tikzpicture}

%% file: fig/environment_coverage/thesis_experiments_coverage_plane_2420f96.tex
% This file was created by matplotlib2tikz v0.7.6.
\begin{tikzpicture}

\definecolor{color0}{rgb}{0.12156862745098,0.227450980392157,0.00392156862745098}

\begin{axis}[
height=\figureheight,
tick align=inside,
tick pos=left,
width=\figurewidth,
x grid style={lightgray!92.0261437908!black},
xmin=-65, xmax=1365,
xtick style={color=black},
y grid style={lightgray!92.0261437908!black},
ylabel={Environment Coverage (cells)},
ymin=-4325.7782908721, ymax=222509.703728137,
ytick style={color=black}
]
% Axis content deleted
% See actual content in: <filename>.tex.bac
\end{axis}

\end{tikzpicture}

%% file: fig/environment_coverage/thesis_experiments_coverage_empty_2420f96.tex
% This file was created by matplotlib2tikz v0.7.6.
\begin{tikzpicture}

\definecolor{color0}{rgb}{0.12156862745098,0.227450980392157,0.00392156862745098}

\begin{axis}[
height=\figureheight,
tick align=inside,
tick pos=left,
width=\figurewidth,
x grid style={lightgray!92.0261437908!black},
xmin=-150, xmax=3150,
xtick style={color=black},
y grid style={lightgray!92.0261437908!black},
ymin=-19786.3786673849, ymax=524750.684698447,
ytick style={color=black}
]
% Axis content deleted
% See actual content in: <filename>.tex.bac
\end{axis}

\end{tikzpicture}

%% file: fig/environment_coverage/thesis_experiments_coverage_indian_tunnel_skylight_2420f96.tex
% This file was created by matplotlib2tikz v0.7.6.
\begin{tikzpicture}

\definecolor{color0}{rgb}{0.12156862745098,0.227450980392157,0.00392156862745098}

\begin{axis}[
height=\figureheight,
tick align=inside,
tick pos=left,
width=\figurewidth,
x grid style={lightgray!92.0261437908!black},
xlabel={Robot-Iteration},
xmin=-110, xmax=2310,
xtick style={color=black},
y grid style={lightgray!92.0261437908!black},
ylabel={Environment Coverage (cells)},
ymin=-6246.20904390939, ymax=230591.343287805,
ytick style={color=black}
]
% Axis content deleted
% See actual content in: <filename>.tex.bac
\end{axis}

\end{tikzpicture}

%% file: fig/environment_coverage/thesis_experiments_coverage_server_room_scaled_0_6_2420f96.tex
% This file was created by matplotlib2tikz v0.7.6.
\begin{tikzpicture}

\definecolor{color0}{rgb}{0.12156862745098,0.227450980392157,0.00392156862745098}

\begin{axis}[
height=\figureheight,
tick align=inside,
tick pos=left,
width=\figurewidth,
x grid style={lightgray!92.0261437908!black},
xlabel={Robot-Iteration},
xmin=-480, xmax=10080,
xtick style={color=black},
y grid style={lightgray!92.0261437908!black},
ymin=-56349.5912806764, ymax=1243373.31387051,
ytick style={color=black}
]
% Axis content deleted
% See actual content in: <filename>.tex.bac
\end{axis}

\end{tikzpicture}

%% file: fig/suboptimality/thesis_experiments_suboptimality_boxes_2420f96.tex
% This file was created by matplotlib2tikz v0.7.6.
\begin{tikzpicture}

\begin{axis}[
height=\figureheight,
legend cell align={left},
legend columns=4,
legend style={at={(0.0,1.05)},anchor=south west, draw=white!80.0!black},
legend image post style={scale=0.5}, % resize legend entries
tick align=inside,
tick pos=left,
width=\figurewidth,
x grid style={lightgray!92.0261437908!black},
xmin=2.6, xmax=33.4,
xtick={4,8,16,32},
xmajorgrids,
xtick style={color=black},
y grid style={lightgray!92.0261437908!black},
ylabel={Suboptimality},
ymajorgrids,
ymin=0.658387038181694, ymax=0.981027569572757,
ytick style={color=black}
]
\path [draw=black, fill=black, opacity=0.2, line width=0.0pt]
(axis cs:4,0.965116017328073)
--(axis cs:8,0.896379488936513)
--(axis cs:16,0.807188122514159)
--(axis cs:32,0.721235218097543)
--(axis cs:32,0.716448979423886)
--(axis cs:16,0.79712554993359)
--(axis cs:8,0.888983192213121)
--(axis cs:4,0.958877032736742)
--cycle;
\path [draw=blue, fill=blue, opacity=0.2, line width=0.0pt]
(axis cs:4,0.953988129804407)
--(axis cs:8,0.873517405085316)
--(axis cs:16,0.765830972566315)
--(axis cs:32,0.680367164951788)
--(axis cs:32,0.673052516881288)
--(axis cs:16,0.753484352792801)
--(axis cs:8,0.860517403297618)
--(axis cs:4,0.948977579349474)
--cycle;
\path [draw=red, fill=red, opacity=0.2, line width=0.0pt]
(axis cs:4,0.966362090873163)
--(axis cs:8,0.893705246448072)
--(axis cs:16,0.791578582002038)
--(axis cs:32,0.710887418281532)
--(axis cs:32,0.707300760888764)
--(axis cs:16,0.784986343223645)
--(axis cs:8,0.885584688064257)
--(axis cs:4,0.95907908486403)
--cycle;
\path [draw=green!50.1960784314!black, fill=green!50.1960784314!black, opacity=0.2, line width=0.0pt]
(axis cs:4,0.956885031013744)
--(axis cs:8,0.899650190421774)
--(axis cs:16,0.803384102939604)
--(axis cs:32,0.71653785062444)
--(axis cs:32,0.711659102755402)
--(axis cs:16,0.794107362470926)
--(axis cs:8,0.892603403139717)
--(axis cs:4,0.953678485574394)
--cycle;
\addplot [semithick, black]
table {%
4 0.961996525032407
8 0.892681340574817
16 0.802156836223875
32 0.718842098760715
};
\addlegendentry{Seq.}
\addplot [semithick, blue]
table {%
4 0.95148285457694
8 0.867017404191467
16 0.759657662679558
32 0.676709840916538
};
\addlegendentry{RSP$_1$}
\addplot [semithick, red]
table {%
4 0.962720587868597
8 0.889644967256164
16 0.788282462612842
32 0.709094089585148
};
\addlegendentry{RSP$_3$}
\addplot [semithick, green!50.1960784314!black]
table {%
4 0.955281758294069
8 0.896126796780746
16 0.798745732705265
32 0.714098476689921
};
\addlegendentry{RSP$_6$}
\end{axis}

\end{tikzpicture}

%% file: fig/suboptimality/thesis_experiments_suboptimality_hallway_2420f96.tex
% This file was created by matplotlib2tikz v0.7.6.
\begin{tikzpicture}

\begin{axis}[
height=\figureheight,
tick align=inside,
tick pos=left,
width=\figurewidth,
x grid style={lightgray!92.0261437908!black},
xmajorgrids,
xmin=2.6, xmax=33.4,
xtick={4,8,16,32},
xtick style={color=black},
y grid style={lightgray!92.0261437908!black},
ymajorgrids,
ymin=0.711726020625745, ymax=0.985500893261604,
ytick style={color=black}
]
\path [draw=black, fill=black, opacity=0.2, line width=0.0pt]
(axis cs:4,0.973056580869064)
--(axis cs:8,0.937612784104185)
--(axis cs:16,0.868929866053388)
--(axis cs:32,0.794605703820135)
--(axis cs:32,0.783155008757146)
--(axis cs:16,0.857586009001771)
--(axis cs:8,0.932424192673601)
--(axis cs:4,0.968062566198584)
--cycle;
\path [draw=blue, fill=blue, opacity=0.2, line width=0.0pt]
(axis cs:4,0.96611232579288)
--(axis cs:8,0.909427040150689)
--(axis cs:16,0.825052525418944)
--(axis cs:32,0.735842722495936)
--(axis cs:32,0.724170333018284)
--(axis cs:16,0.81614416259534)
--(axis cs:8,0.900504726436803)
--(axis cs:4,0.957151754516616)
--cycle;
\path [draw=red, fill=red, opacity=0.2, line width=0.0pt]
(axis cs:4,0.972924808120328)
--(axis cs:8,0.93037849979574)
--(axis cs:16,0.86213362806924)
--(axis cs:32,0.789618941579381)
--(axis cs:32,0.780068252263833)
--(axis cs:16,0.852447813973787)
--(axis cs:8,0.924063533123661)
--(axis cs:4,0.968175091746736)
--cycle;
\path [draw=green!50.1960784314!black, fill=green!50.1960784314!black, opacity=0.2, line width=0.0pt]
(axis cs:4,0.971553613974181)
--(axis cs:8,0.934229092551815)
--(axis cs:16,0.867806284724248)
--(axis cs:32,0.797554722937503)
--(axis cs:32,0.792800583670256)
--(axis cs:16,0.859596181505249)
--(axis cs:8,0.92697046604468)
--(axis cs:4,0.966975969192826)
--cycle;
\addplot [semithick, black]
table {%
4 0.970559573533824
8 0.935018488388893
16 0.86325793752758
32 0.788880356288641
};
\addplot [semithick, blue]
table {%
4 0.961632040154748
8 0.904965883293746
16 0.820598344007142
32 0.73000652775711
};
\addplot [semithick, red]
table {%
4 0.970549949933532
8 0.9272210164597
16 0.857290721021513
32 0.784843596921607
};
\addplot [semithick, green!50.1960784314!black]
table {%
4 0.969264791583504
8 0.930599779298248
16 0.863701233114749
32 0.795177653303879
};
\end{axis}

\end{tikzpicture}

%% file: fig/suboptimality/thesis_experiments_suboptimality_plane_2420f96.tex
% This file was created by matplotlib2tikz v0.7.6.
\begin{tikzpicture}

\begin{axis}[
height=\figureheight,
tick align=inside,
tick pos=left,
width=\figurewidth,
x grid style={lightgray!92.0261437908!black},
xmajorgrids,
xmin=2.6, xmax=33.4,
xtick={4,8,16,32},
xtick style={color=black},
y grid style={lightgray!92.0261437908!black},
ymajorgrids,
ymin=0.727006781683169, ymax=0.998007210509476,
ytick style={color=black}
]
\path [draw=black, fill=black, opacity=0.2, line width=0.0pt]
(axis cs:4,0.985689009199189)
--(axis cs:8,0.95865769440926)
--(axis cs:16,0.901367414800235)
--(axis cs:32,0.798446560293545)
--(axis cs:32,0.788607291762268)
--(axis cs:16,0.898722483428229)
--(axis cs:8,0.953017872723794)
--(axis cs:4,0.981980224589827)
--cycle;
\path [draw=blue, fill=blue, opacity=0.2, line width=0.0pt]
(axis cs:4,0.977865516086646)
--(axis cs:8,0.944937830645764)
--(axis cs:16,0.862403323377394)
--(axis cs:32,0.748903629279214)
--(axis cs:32,0.739324982993456)
--(axis cs:16,0.854609188214293)
--(axis cs:8,0.941125780246909)
--(axis cs:4,0.971692045717908)
--cycle;
\path [draw=red, fill=red, opacity=0.2, line width=0.0pt]
(axis cs:4,0.983883503499852)
--(axis cs:8,0.957750679741461)
--(axis cs:16,0.897107667036728)
--(axis cs:32,0.786800945539797)
--(axis cs:32,0.77977416445793)
--(axis cs:16,0.887720886623516)
--(axis cs:8,0.953483884838492)
--(axis cs:4,0.981049282856595)
--cycle;
\path [draw=green!50.1960784314!black, fill=green!50.1960784314!black, opacity=0.2, line width=0.0pt]
(axis cs:4,0.980440698370675)
--(axis cs:8,0.95829694912201)
--(axis cs:16,0.895794213965701)
--(axis cs:32,0.785304109354172)
--(axis cs:32,0.779232285519457)
--(axis cs:16,0.891097506441789)
--(axis cs:8,0.955640077682401)
--(axis cs:4,0.976232078671403)
--cycle;
\addplot [semithick, black]
table {%
4 0.983834616894508
8 0.955837783566527
16 0.900044949114232
32 0.793526926027906
};
\addplot [semithick, blue]
table {%
4 0.974778780902277
8 0.943031805446336
16 0.858506255795844
32 0.744114306136335
};
\addplot [semithick, red]
table {%
4 0.982466393178223
8 0.955617282289976
16 0.892414276830122
32 0.783287554998863
};
\addplot [semithick, green!50.1960784314!black]
table {%
4 0.978336388521039
8 0.956968513402205
16 0.893445860203745
32 0.782268197436814
};
\end{axis}

\end{tikzpicture}

%% file: fig/suboptimality/thesis_experiments_suboptimality_empty_2420f96.tex
% This file was created by matplotlib2tikz v0.7.6.
\begin{tikzpicture}

\begin{axis}[
height=\figureheight,
tick align=inside,
tick pos=left,
width=\figurewidth,
x grid style={lightgray!92.0261437908!black},
xlabel={Num. Robots},
xmajorgrids,
xmin=2.6, xmax=33.4,
xtick={4,8,16,32},
xtick style={color=black},
y grid style={lightgray!92.0261437908!black},
ylabel={Suboptimality},
ymajorgrids,
ymin=0.692562618329813, ymax=0.999455380917714,
ytick style={color=black}
]
\path [draw=black, fill=black, opacity=0.2, line width=0.0pt]
(axis cs:4,0.985505709890991)
--(axis cs:8,0.951945491318529)
--(axis cs:16,0.87922065846736)
--(axis cs:32,0.800895602577976)
--(axis cs:32,0.788247770024664)
--(axis cs:16,0.871785444117147)
--(axis cs:8,0.946896644506391)
--(axis cs:4,0.983465038486744)
--cycle;
\path [draw=blue, fill=blue, opacity=0.2, line width=0.0pt]
(axis cs:4,0.97105818144169)
--(axis cs:8,0.925816121304062)
--(axis cs:16,0.828404329560753)
--(axis cs:32,0.722652725561558)
--(axis cs:32,0.706512289356536)
--(axis cs:16,0.813188891503177)
--(axis cs:8,0.91792655533893)
--(axis cs:4,0.963665313116637)
--cycle;
\path [draw=red, fill=red, opacity=0.2, line width=0.0pt]
(axis cs:4,0.977135088166194)
--(axis cs:8,0.950432468700303)
--(axis cs:16,0.886860830489869)
--(axis cs:32,0.81094573838937)
--(axis cs:32,0.796467303209899)
--(axis cs:16,0.879256225158327)
--(axis cs:8,0.945528631812611)
--(axis cs:4,0.972344772490698)
--cycle;
\path [draw=green!50.1960784314!black, fill=green!50.1960784314!black, opacity=0.2, line width=0.0pt]
(axis cs:4,0.980764852112999)
--(axis cs:8,0.946827673858836)
--(axis cs:16,0.885570216982605)
--(axis cs:32,0.80033510898552)
--(axis cs:32,0.786229639679025)
--(axis cs:16,0.879026388502622)
--(axis cs:8,0.942154493935499)
--(axis cs:4,0.976898922780705)
--cycle;
\addplot [semithick, black]
table {%
4 0.984485374188867
8 0.94942106791246
16 0.875503051292254
32 0.79457168630132
};
\addplot [semithick, blue]
table {%
4 0.967361747279164
8 0.921871338321496
16 0.820796610531965
32 0.714582507459047
};
\addplot [semithick, red]
table {%
4 0.974739930328446
8 0.947980550256457
16 0.883058527824098
32 0.803706520799635
};
\addplot [semithick, green!50.1960784314!black]
table {%
4 0.978831887446852
8 0.944491083897168
16 0.882298302742614
32 0.793282374332272
};
\end{axis}

\end{tikzpicture}

%% file: fig/suboptimality/thesis_experiments_suboptimality_indian_tunnel_skylight_2420f96.tex
% This file was created by matplotlib2tikz v0.7.6.
\begin{tikzpicture}

\begin{axis}[
height=\figureheight,
tick align=inside,
tick pos=left,
width=\figurewidth,
x grid style={lightgray!92.0261437908!black},
xlabel={Num. Robots},
xmajorgrids,
xmin=2.6, xmax=33.4,
xtick={4,8,16,32},
xtick style={color=black},
y grid style={lightgray!92.0261437908!black},
ymajorgrids,
ymin=0.761482608230704, ymax=0.999886989203023,
ytick style={color=black}
]
\path [draw=black, fill=black, opacity=0.2, line width=0.0pt]
(axis cs:4,0.989050426431554)
--(axis cs:8,0.963642695230361)
--(axis cs:16,0.908591872466488)
--(axis cs:32,0.832681730457248)
--(axis cs:32,0.826728390200297)
--(axis cs:16,0.899680368339212)
--(axis cs:8,0.958744716434054)
--(axis cs:4,0.986704010310967)
--cycle;
\path [draw=blue, fill=blue, opacity=0.2, line width=0.0pt]
(axis cs:4,0.986849812167453)
--(axis cs:8,0.95316283032738)
--(axis cs:16,0.876396278414805)
--(axis cs:32,0.783418124247361)
--(axis cs:32,0.772319171002173)
--(axis cs:16,0.865011950394842)
--(axis cs:8,0.946511701959897)
--(axis cs:4,0.984082638431111)
--cycle;
\path [draw=red, fill=red, opacity=0.2, line width=0.0pt]
(axis cs:4,0.988741682368799)
--(axis cs:8,0.964232421702813)
--(axis cs:16,0.904385639129174)
--(axis cs:32,0.823449947254179)
--(axis cs:32,0.815733815700559)
--(axis cs:16,0.897201495402442)
--(axis cs:8,0.959069299106235)
--(axis cs:4,0.985939777795143)
--cycle;
\path [draw=green!50.1960784314!black, fill=green!50.1960784314!black, opacity=0.2, line width=0.0pt]
(axis cs:4,0.988821458332432)
--(axis cs:8,0.964038578558087)
--(axis cs:16,0.908351874531167)
--(axis cs:32,0.827811391311039)
--(axis cs:32,0.820359776643949)
--(axis cs:16,0.904110433589344)
--(axis cs:8,0.96009649694951)
--(axis cs:4,0.985567868055238)
--cycle;
\addplot [semithick, black]
table {%
4 0.987877218371261
8 0.961193705832208
16 0.90413612040285
32 0.829705060328773
};
\addplot [semithick, blue]
table {%
4 0.985466225299282
8 0.949837266143639
16 0.870704114404823
32 0.777868647624767
};
\addplot [semithick, red]
table {%
4 0.987340730081971
8 0.961650860404524
16 0.900793567265808
32 0.819591881477369
};
\addplot [semithick, green!50.1960784314!black]
table {%
4 0.987194663193835
8 0.962067537753798
16 0.906231154060255
32 0.824085583977494
};
\end{axis}

\end{tikzpicture}

%% file: fig/suboptimality/thesis_experiments_suboptimality_server_room_scaled_0_6_2420f96.tex
% This file was created by matplotlib2tikz v0.7.6.
\begin{tikzpicture}

\begin{axis}[
height=\figureheight,
tick align=inside,
tick pos=left,
width=\figurewidth,
x grid style={lightgray!92.0261437908!black},
xlabel={Num. Robots},
xmajorgrids,
xmin=2.6, xmax=33.4,
xtick={4,8,16,32},
xtick style={color=black},
y grid style={lightgray!92.0261437908!black},
ymajorgrids,
ymin=0.844703904258497, ymax=1.00040477604721,
ytick style={color=black}
]
\path [draw=black, fill=black, opacity=0.2, line width=0.0pt]
(axis cs:4,0.991739582588112)
--(axis cs:8,0.976005205461024)
--(axis cs:16,0.935121872854867)
--(axis cs:32,0.867507433009072)
--(axis cs:32,0.862327617942758)
--(axis cs:16,0.931128616926046)
--(axis cs:8,0.971739737261198)
--(axis cs:4,0.99081701279236)
--cycle;
\path [draw=green!50.1960784314!black, fill=green!50.1960784314!black, opacity=0.2, line width=0.0pt]
(axis cs:4,0.993327463693177)
--(axis cs:8,0.975570604782392)
--(axis cs:16,0.930124613447741)
--(axis cs:32,0.856450672558353)
--(axis cs:32,0.851781216612529)
--(axis cs:16,0.924512089604038)
--(axis cs:8,0.972720360316847)
--(axis cs:4,0.99219533172419)
--cycle;
\addplot [semithick, black]
table {%
4 0.991278297690236
8 0.973872471361111
16 0.933125244890456
32 0.864917525475915
};
\addplot [semithick, green!50.1960784314!black]
table {%
4 0.992761397708684
8 0.97414548254962
16 0.927318351525889
32 0.854115944585441
};
\end{axis}

\end{tikzpicture}

%% file: fig/test_csqmi_coverage_coverage_boxes_afdb01c.tex
% This file was created by matplotlib2tikz v0.7.6.
\begin{tikzpicture}

\begin{axis}[
height=\figureheight,
legend cell align={left},
legend style={at={(0.97,0.03)}, anchor=south east, draw=white!80.0!black},
tick align=outside,
tick pos=left,
width=\figurewidth,
x grid style={lightgray!92.0261437908!black},
xlabel={Robot-Iteration},
xmin=-100, xmax=2100,
xtick style={color=black},
y grid style={lightgray!92.0261437908!black},
ylabel={Environment Coverage (cells)},
ymin=1252.93614457773, ymax=209880.050579726,
ytick style={color=black}
]
\path [draw=blue, fill=blue, opacity=0.2, line width=0.0pt]
(axis cs:0,11791.6131992792)
--(axis cs:16,12814.4587842683)
--(axis cs:32,16370.2161507199)
--(axis cs:48,20610.5127904066)
--(axis cs:64,23967.2661752056)
--(axis cs:80,27761.8988248936)
--(axis cs:96,31683.7187086181)
--(axis cs:112,35428.9052663621)
--(axis cs:128,38067.5112653852)
--(axis cs:144,42461.5552971157)
--(axis cs:160,46449.302501768)
--(axis cs:176,50046.569792839)
--(axis cs:192,52036.0280951828)
--(axis cs:208,55267.9990676392)
--(axis cs:224,59586.193541516)
--(axis cs:240,62841.3087556591)
--(axis cs:256,66410.7428761976)
--(axis cs:272,70468.07031332)
--(axis cs:288,73635.6362641868)
--(axis cs:304,78842.3205157824)
--(axis cs:320,82348.4068689743)
--(axis cs:336,87362.459658818)
--(axis cs:352,89799.6467855939)
--(axis cs:368,93195.1113285149)
--(axis cs:384,96264.7293779839)
--(axis cs:400,100493.83065729)
--(axis cs:416,104454.380059259)
--(axis cs:432,105567.173197527)
--(axis cs:448,109847.917501212)
--(axis cs:464,112342.052273705)
--(axis cs:480,115387.143295704)
--(axis cs:496,117614.342748)
--(axis cs:512,119587.886518511)
--(axis cs:528,122459.899782865)
--(axis cs:544,123688.340942527)
--(axis cs:560,127302.838707134)
--(axis cs:576,129305.241497129)
--(axis cs:592,131035.645572088)
--(axis cs:608,133446.360258235)
--(axis cs:624,136070.738208583)
--(axis cs:640,138927.152966785)
--(axis cs:656,140945.78937897)
--(axis cs:672,143999.269761375)
--(axis cs:688,146172.925590663)
--(axis cs:704,148627.943092213)
--(axis cs:720,152340.072089388)
--(axis cs:736,155386.725481456)
--(axis cs:752,158196.205399737)
--(axis cs:768,161184.432020907)
--(axis cs:784,163896.93158628)
--(axis cs:800,166089.462627878)
--(axis cs:816,169120.522632861)
--(axis cs:832,171597.681793476)
--(axis cs:848,175016.835371284)
--(axis cs:864,176388.645373075)
--(axis cs:880,179490.437552302)
--(axis cs:896,181636.027251594)
--(axis cs:912,183759.871096222)
--(axis cs:928,185036.533975243)
--(axis cs:944,186145.54180503)
--(axis cs:960,187385.330733329)
--(axis cs:976,188766.917364086)
--(axis cs:992,189643.351625154)
--(axis cs:1008,189979.300616721)
--(axis cs:1024,190717.414153404)
--(axis cs:1040,191405.005505477)
--(axis cs:1056,191878.803770545)
--(axis cs:1072,192195.727173761)
--(axis cs:1088,192389.343638083)
--(axis cs:1104,192735.797869098)
--(axis cs:1120,192975.793588611)
--(axis cs:1136,193262.35525741)
--(axis cs:1152,193492.827324956)
--(axis cs:1168,193885.564292513)
--(axis cs:1184,194069.175810618)
--(axis cs:1200,194299.576585007)
--(axis cs:1216,194511.638026518)
--(axis cs:1232,194729.609424043)
--(axis cs:1248,194867.180619797)
--(axis cs:1264,195157.809257421)
--(axis cs:1280,195345.481178986)
--(axis cs:1296,195477.22663977)
--(axis cs:1312,195720.318583221)
--(axis cs:1328,195907.331525902)
--(axis cs:1344,196074.142188223)
--(axis cs:1360,196248.080334154)
--(axis cs:1376,196345.63929744)
--(axis cs:1392,196504.625737978)
--(axis cs:1408,196706.832873917)
--(axis cs:1424,196857.910028882)
--(axis cs:1440,196951.221631627)
--(axis cs:1456,197087.253656869)
--(axis cs:1472,197213.979675759)
--(axis cs:1488,197362.295552653)
--(axis cs:1504,197487.977893921)
--(axis cs:1520,197640.920347373)
--(axis cs:1536,197728.623596293)
--(axis cs:1552,197850.149944702)
--(axis cs:1568,197960.63938945)
--(axis cs:1584,198067.403476792)
--(axis cs:1600,198231.053048499)
--(axis cs:1616,198355.401615404)
--(axis cs:1632,198491.136855912)
--(axis cs:1648,198604.386430576)
--(axis cs:1664,198711.514216363)
--(axis cs:1680,198872.790221752)
--(axis cs:1696,198971.906644565)
--(axis cs:1712,199085.480562331)
--(axis cs:1728,199227.54694209)
--(axis cs:1744,199328.701962645)
--(axis cs:1760,199424.406168599)
--(axis cs:1776,199496.136982739)
--(axis cs:1792,199613.895235144)
--(axis cs:1808,199661.583689714)
--(axis cs:1824,199749.226580583)
--(axis cs:1840,199827.483682828)
--(axis cs:1856,199902.197490854)
--(axis cs:1872,199981.194145917)
--(axis cs:1888,200064.225455829)
--(axis cs:1904,200133.785922079)
--(axis cs:1920,200181.46119587)
--(axis cs:1936,200230.141061066)
--(axis cs:1952,200288.852667426)
--(axis cs:1968,200329.593387117)
--(axis cs:1984,200358.576573613)
--(axis cs:2000,200396.999923582)
--(axis cs:2000,200106.200076418)
--(axis cs:1984,200051.823426387)
--(axis cs:1968,200016.406612883)
--(axis cs:1952,199961.547332574)
--(axis cs:1936,199898.858938934)
--(axis cs:1920,199819.13880413)
--(axis cs:1904,199774.214077921)
--(axis cs:1888,199687.774544171)
--(axis cs:1872,199594.005854083)
--(axis cs:1856,199482.402509146)
--(axis cs:1840,199364.116317172)
--(axis cs:1824,199265.973419417)
--(axis cs:1808,199197.816310286)
--(axis cs:1792,199122.104764856)
--(axis cs:1776,199013.863017261)
--(axis cs:1760,198949.593831401)
--(axis cs:1744,198874.098037355)
--(axis cs:1728,198763.45305791)
--(axis cs:1712,198567.319437669)
--(axis cs:1696,198416.893355435)
--(axis cs:1680,198291.009778248)
--(axis cs:1664,198092.885783637)
--(axis cs:1648,197959.813569424)
--(axis cs:1632,197831.263144088)
--(axis cs:1616,197711.598384596)
--(axis cs:1600,197525.146951501)
--(axis cs:1584,197401.396523208)
--(axis cs:1568,197299.96061055)
--(axis cs:1552,197182.050055298)
--(axis cs:1536,197072.376403707)
--(axis cs:1520,196978.679652627)
--(axis cs:1504,196803.422106079)
--(axis cs:1488,196666.504447347)
--(axis cs:1472,196516.420324241)
--(axis cs:1456,196400.946343132)
--(axis cs:1440,196265.578368373)
--(axis cs:1424,196168.889971118)
--(axis cs:1408,195993.367126083)
--(axis cs:1392,195795.574262022)
--(axis cs:1376,195590.56070256)
--(axis cs:1360,195501.119665846)
--(axis cs:1344,195270.057811777)
--(axis cs:1328,195095.868474098)
--(axis cs:1312,194769.681416779)
--(axis cs:1296,194492.37336023)
--(axis cs:1280,194358.718821014)
--(axis cs:1264,194103.990742579)
--(axis cs:1248,193818.019380203)
--(axis cs:1232,193684.990575957)
--(axis cs:1216,193464.961973482)
--(axis cs:1200,193196.223414994)
--(axis cs:1184,193013.824189382)
--(axis cs:1168,192789.035707487)
--(axis cs:1152,192443.372675044)
--(axis cs:1136,192212.244742591)
--(axis cs:1120,191882.606411389)
--(axis cs:1104,191655.402130902)
--(axis cs:1088,191252.856361917)
--(axis cs:1072,190995.072826239)
--(axis cs:1056,190715.396229455)
--(axis cs:1040,190339.394494523)
--(axis cs:1024,189520.985846596)
--(axis cs:1008,188662.299383279)
--(axis cs:992,188313.048374846)
--(axis cs:976,187000.482635914)
--(axis cs:960,185718.669266671)
--(axis cs:944,184511.85819497)
--(axis cs:928,182426.266024757)
--(axis cs:912,181211.328903778)
--(axis cs:896,178044.772748406)
--(axis cs:880,175847.362447699)
--(axis cs:864,172356.154626926)
--(axis cs:848,170750.564628716)
--(axis cs:832,166927.518206524)
--(axis cs:816,164584.077367139)
--(axis cs:800,161514.937372122)
--(axis cs:784,159096.46841372)
--(axis cs:768,156820.167979093)
--(axis cs:752,153125.994600263)
--(axis cs:736,150354.674518545)
--(axis cs:720,147541.327910612)
--(axis cs:704,145117.056907787)
--(axis cs:688,142577.274409337)
--(axis cs:672,140147.530238625)
--(axis cs:656,137699.41062103)
--(axis cs:640,135920.247033215)
--(axis cs:624,133118.861791417)
--(axis cs:608,130798.639741765)
--(axis cs:592,128822.354427912)
--(axis cs:576,126820.958502871)
--(axis cs:560,124684.161292866)
--(axis cs:544,121620.659057473)
--(axis cs:528,120100.900217135)
--(axis cs:512,117098.313481489)
--(axis cs:496,115176.057252)
--(axis cs:480,113146.256704296)
--(axis cs:464,109507.347726295)
--(axis cs:448,106847.282498788)
--(axis cs:432,102754.026802473)
--(axis cs:416,101828.619940741)
--(axis cs:400,97302.9693427096)
--(axis cs:384,93207.2706220161)
--(axis cs:368,90696.4886714851)
--(axis cs:352,87702.7532144061)
--(axis cs:336,83989.740341182)
--(axis cs:320,79701.3931310257)
--(axis cs:304,76300.6794842176)
--(axis cs:288,71513.7637358132)
--(axis cs:272,68875.12968668)
--(axis cs:256,64749.4571238024)
--(axis cs:240,60920.8912443409)
--(axis cs:224,58203.006458484)
--(axis cs:208,53742.6009323608)
--(axis cs:192,50393.1719048173)
--(axis cs:176,48218.630207161)
--(axis cs:160,44410.2974982321)
--(axis cs:144,40407.4447028843)
--(axis cs:128,36678.2887346148)
--(axis cs:112,33703.6947336379)
--(axis cs:96,30074.2812913819)
--(axis cs:80,26181.9011751064)
--(axis cs:64,22520.3338247944)
--(axis cs:48,19426.8872095934)
--(axis cs:32,15307.9838492801)
--(axis cs:16,11734.1412157317)
--(axis cs:0,10735.9868007208)
--cycle;
\path [draw=red, fill=red, opacity=0.2, line width=0.0pt]
(axis cs:0,11922.3991499688)
--(axis cs:16,13220.6965227383)
--(axis cs:32,15615.7675797111)
--(axis cs:48,18833.9684376873)
--(axis cs:64,23797.0817077075)
--(axis cs:80,26605.254293114)
--(axis cs:96,29381.4611558966)
--(axis cs:112,32714.0981684197)
--(axis cs:128,36837.8641364311)
--(axis cs:144,39130.8344839425)
--(axis cs:160,42248.5934068463)
--(axis cs:176,45397.2085485096)
--(axis cs:192,48875.8889174792)
--(axis cs:208,52198.9744904567)
--(axis cs:224,53889.1075329336)
--(axis cs:240,57193.6985055578)
--(axis cs:256,59979.788883484)
--(axis cs:272,63734.5128552983)
--(axis cs:288,66382.3093771809)
--(axis cs:304,69860.8260068673)
--(axis cs:320,71839.1927087545)
--(axis cs:336,75162.4433519879)
--(axis cs:352,78198.6993945579)
--(axis cs:368,81767.1831769685)
--(axis cs:384,85321.0174469294)
--(axis cs:400,87441.998757681)
--(axis cs:416,90841.7257644284)
--(axis cs:432,93863.0276484236)
--(axis cs:448,97247.5856053652)
--(axis cs:464,99435.4367736241)
--(axis cs:480,101938.382721083)
--(axis cs:496,105761.123299109)
--(axis cs:512,106442.992762161)
--(axis cs:528,109419.403044816)
--(axis cs:544,111464.023617291)
--(axis cs:560,113277.150201318)
--(axis cs:576,115867.063071371)
--(axis cs:592,117523.169039423)
--(axis cs:608,119822.023346253)
--(axis cs:624,122219.018372283)
--(axis cs:640,123717.346445785)
--(axis cs:656,125526.039160012)
--(axis cs:672,127716.590011619)
--(axis cs:688,129868.485595115)
--(axis cs:704,131869.039873577)
--(axis cs:720,133277.535066935)
--(axis cs:736,135742.411192985)
--(axis cs:752,137908.418381452)
--(axis cs:768,139215.480206174)
--(axis cs:784,141551.221886375)
--(axis cs:800,144056.741859573)
--(axis cs:816,146031.421777798)
--(axis cs:832,149269.512227657)
--(axis cs:848,151109.101472137)
--(axis cs:864,153608.968312951)
--(axis cs:880,156716.70243012)
--(axis cs:896,159356.382412629)
--(axis cs:912,160299.904644111)
--(axis cs:928,163053.223778621)
--(axis cs:944,166033.36337532)
--(axis cs:960,168486.689351151)
--(axis cs:976,170042.456431183)
--(axis cs:992,172400.132455017)
--(axis cs:1008,173966.310381262)
--(axis cs:1024,176759.626621588)
--(axis cs:1040,178645.666745889)
--(axis cs:1056,180159.002167246)
--(axis cs:1072,181395.612629663)
--(axis cs:1088,182027.215719382)
--(axis cs:1104,183153.903519238)
--(axis cs:1120,183569.844268013)
--(axis cs:1136,184459.643381211)
--(axis cs:1152,185024.611278622)
--(axis cs:1168,185577.163123591)
--(axis cs:1184,185910.494270395)
--(axis cs:1200,186477.249479095)
--(axis cs:1216,186787.368237444)
--(axis cs:1232,187029.482456116)
--(axis cs:1248,187414.896501201)
--(axis cs:1264,187869.827705498)
--(axis cs:1280,188248.467900678)
--(axis cs:1296,188803.282962173)
--(axis cs:1312,189082.836475319)
--(axis cs:1328,189371.183557451)
--(axis cs:1344,189633.755876401)
--(axis cs:1360,189751.909587686)
--(axis cs:1376,189950.387973918)
--(axis cs:1392,190354.36007323)
--(axis cs:1408,190509.882370453)
--(axis cs:1424,190818.955330431)
--(axis cs:1440,191165.949171022)
--(axis cs:1456,191428.784841883)
--(axis cs:1472,191667.162669574)
--(axis cs:1488,191839.194013285)
--(axis cs:1504,191999.876115925)
--(axis cs:1520,192157.547781939)
--(axis cs:1536,192341.702371654)
--(axis cs:1552,192511.632120065)
--(axis cs:1568,192652.049248371)
--(axis cs:1584,192927.433167745)
--(axis cs:1600,193043.036942923)
--(axis cs:1616,193135.355197005)
--(axis cs:1632,193235.007117886)
--(axis cs:1648,193383.464736589)
--(axis cs:1664,193456.686578441)
--(axis cs:1680,193557.723336685)
--(axis cs:1696,193654.437700829)
--(axis cs:1712,193770.763136083)
--(axis cs:1728,193833.08662054)
--(axis cs:1744,193927.720372173)
--(axis cs:1760,194001.379195791)
--(axis cs:1776,194062.325573372)
--(axis cs:1792,194205.026840451)
--(axis cs:1808,194285.436230677)
--(axis cs:1824,194432.776778111)
--(axis cs:1840,194525.020693858)
--(axis cs:1856,194587.963338236)
--(axis cs:1872,194693.873647086)
--(axis cs:1888,194819.458394578)
--(axis cs:1904,195028.836923159)
--(axis cs:1920,195118.557995677)
--(axis cs:1936,195227.06946458)
--(axis cs:1952,195378.678100046)
--(axis cs:1968,195476.260467964)
--(axis cs:1984,195546.39221075)
--(axis cs:2000,195639.835544741)
--(axis cs:2000,194783.164455259)
--(axis cs:1984,194688.80778925)
--(axis cs:1968,194618.739532036)
--(axis cs:1952,194507.921899954)
--(axis cs:1936,194362.93053542)
--(axis cs:1920,194206.842004323)
--(axis cs:1904,194145.163076841)
--(axis cs:1888,193895.341605422)
--(axis cs:1872,193778.926352914)
--(axis cs:1856,193683.636661764)
--(axis cs:1840,193605.779306142)
--(axis cs:1824,193516.423221889)
--(axis cs:1808,193357.763769323)
--(axis cs:1792,193232.173159549)
--(axis cs:1776,193088.674426628)
--(axis cs:1760,192986.220804209)
--(axis cs:1744,192925.079627827)
--(axis cs:1728,192812.71337946)
--(axis cs:1712,192700.436863917)
--(axis cs:1696,192567.362299171)
--(axis cs:1680,192460.476663315)
--(axis cs:1664,192374.113421559)
--(axis cs:1648,192259.735263411)
--(axis cs:1632,192114.192882114)
--(axis cs:1616,192014.244802995)
--(axis cs:1600,191926.563057077)
--(axis cs:1584,191753.166832255)
--(axis cs:1568,191531.150751629)
--(axis cs:1552,191400.767879935)
--(axis cs:1536,191180.297628346)
--(axis cs:1520,190997.252218061)
--(axis cs:1504,190848.923884075)
--(axis cs:1488,190707.805986715)
--(axis cs:1472,190556.637330426)
--(axis cs:1456,190353.615158117)
--(axis cs:1440,189960.650828978)
--(axis cs:1424,189763.844669569)
--(axis cs:1408,189434.117629547)
--(axis cs:1392,189308.03992677)
--(axis cs:1376,188898.612026082)
--(axis cs:1360,188751.890412314)
--(axis cs:1344,188614.844123599)
--(axis cs:1328,188435.416442549)
--(axis cs:1312,188175.763524681)
--(axis cs:1296,187967.317037827)
--(axis cs:1280,187506.532099322)
--(axis cs:1264,187290.372294502)
--(axis cs:1248,186823.903498799)
--(axis cs:1232,186393.317543884)
--(axis cs:1216,186150.031762556)
--(axis cs:1200,185749.350520905)
--(axis cs:1184,185201.705729605)
--(axis cs:1168,184838.636876409)
--(axis cs:1152,184070.188721378)
--(axis cs:1136,183412.756618789)
--(axis cs:1120,182173.155731987)
--(axis cs:1104,181473.296480762)
--(axis cs:1088,179839.384280618)
--(axis cs:1072,179065.987370337)
--(axis cs:1056,177636.597832754)
--(axis cs:1040,175560.133254111)
--(axis cs:1024,173805.773378412)
--(axis cs:1008,170868.089618738)
--(axis cs:992,169014.867544983)
--(axis cs:976,167074.143568817)
--(axis cs:960,165417.510648849)
--(axis cs:944,162709.83662468)
--(axis cs:928,160086.576221379)
--(axis cs:912,157500.895355889)
--(axis cs:896,156188.217587371)
--(axis cs:880,153675.69756988)
--(axis cs:864,150719.631687049)
--(axis cs:848,147882.498527863)
--(axis cs:832,146721.087772343)
--(axis cs:816,143100.178222202)
--(axis cs:800,141309.658140427)
--(axis cs:784,138955.978113625)
--(axis cs:768,136814.919793826)
--(axis cs:752,135446.981618548)
--(axis cs:736,133083.988807015)
--(axis cs:720,131431.664933065)
--(axis cs:704,129318.160126423)
--(axis cs:688,127542.514404885)
--(axis cs:672,125984.809988381)
--(axis cs:656,123408.760839988)
--(axis cs:640,121712.453554215)
--(axis cs:624,120165.181627717)
--(axis cs:608,117576.576653747)
--(axis cs:592,114907.830960577)
--(axis cs:576,113020.136928629)
--(axis cs:560,110607.249798682)
--(axis cs:544,108587.576382709)
--(axis cs:528,106552.596955184)
--(axis cs:512,103280.607237839)
--(axis cs:496,102751.876700891)
--(axis cs:480,98662.2172789169)
--(axis cs:464,95600.1632263759)
--(axis cs:448,93754.0143946348)
--(axis cs:432,90936.5723515765)
--(axis cs:416,87798.8742355716)
--(axis cs:400,84097.001242319)
--(axis cs:384,82832.1825530706)
--(axis cs:368,79363.6168230315)
--(axis cs:352,75488.1006054422)
--(axis cs:336,72642.3566480121)
--(axis cs:320,69327.2072912455)
--(axis cs:304,67215.1739931327)
--(axis cs:288,63747.8906228191)
--(axis cs:272,61269.8871447018)
--(axis cs:256,57901.211116516)
--(axis cs:240,55198.9014944422)
--(axis cs:224,52246.0924670664)
--(axis cs:208,50266.8255095433)
--(axis cs:192,46975.9110825208)
--(axis cs:176,43623.7914514904)
--(axis cs:160,40296.4065931537)
--(axis cs:144,37416.7655160575)
--(axis cs:128,35525.7358635689)
--(axis cs:112,31270.9018315803)
--(axis cs:96,28149.9388441034)
--(axis cs:80,25064.345706886)
--(axis cs:64,22641.1182922925)
--(axis cs:48,17923.6315623127)
--(axis cs:32,14456.0324202889)
--(axis cs:16,12391.5034772617)
--(axis cs:0,11080.6008500312)
--cycle;
\addplot [semithick, blue]
table {%
0 11263.8
16 12274.3
32 15839.1
48 20018.7
64 23243.8
80 26971.9
96 30879
112 34566.3
128 37372.9
144 41434.5
160 45429.8
176 49132.6
192 51214.6
208 54505.3
224 58894.6
240 61881.1
256 65580.1
272 69671.6
288 72574.7
304 77571.5
320 81024.9
336 85676.1
352 88751.2
368 91945.8
384 94736
400 98898.4
416 103141.5
432 104160.6
448 108347.6
464 110924.7
480 114266.7
496 116395.2
512 118343.1
528 121280.4
544 122654.5
560 125993.5
576 128063.1
592 129929
608 132122.5
624 134594.8
640 137423.7
656 139322.6
672 142073.4
688 144375.1
704 146872.5
720 149940.7
736 152870.7
752 155661.1
768 159002.3
784 161496.7
800 163802.2
816 166852.3
832 169262.6
848 172883.7
864 174372.4
880 177668.9
896 179840.4
912 182485.6
928 183731.4
944 185328.7
960 186552
976 187883.7
992 188978.2
1008 189320.8
1024 190119.2
1040 190872.2
1056 191297.1
1072 191595.4
1088 191821.1
1104 192195.6
1120 192429.2
1136 192737.3
1152 192968.1
1168 193337.3
1184 193541.5
1200 193747.9
1216 193988.3
1232 194207.3
1248 194342.6
1264 194630.9
1280 194852.1
1296 194984.8
1312 195245
1328 195501.6
1344 195672.1
1360 195874.6
1376 195968.1
1392 196150.1
1408 196350.1
1424 196513.4
1440 196608.4
1456 196744.1
1472 196865.2
1488 197014.4
1504 197145.7
1520 197309.8
1536 197400.5
1552 197516.1
1568 197630.3
1584 197734.4
1600 197878.1
1616 198033.5
1632 198161.2
1648 198282.1
1664 198402.2
1680 198581.9
1696 198694.4
1712 198826.4
1728 198995.5
1744 199101.4
1760 199187
1776 199255
1792 199368
1808 199429.7
1824 199507.6
1840 199595.8
1856 199692.3
1872 199787.6
1888 199876
1904 199954
1920 200000.3
1936 200064.5
1952 200125.2
1968 200173
1984 200205.2
2000 200251.6
};
\addlegendentry{Optimistic Coverage}
\addplot [semithick, red]
table {%
0 11501.5
16 12806.1
32 15035.9
48 18378.8
64 23219.1
80 25834.8
96 28765.7
112 31992.5
128 36181.8
144 38273.8
160 41272.5
176 44510.5
192 47925.9
208 51232.9
224 53067.6
240 56196.3
256 58940.5
272 62502.2
288 65065.1
304 68538
320 70583.2
336 73902.4
352 76843.4
368 80565.4
384 84076.6
400 85769.5
416 89320.3
432 92399.8
448 95500.8
464 97517.8
480 100300.3
496 104256.5
512 104861.8
528 107986
544 110025.8
560 111942.2
576 114443.6
592 116215.5
608 118699.3
624 121192.1
640 122714.9
656 124467.4
672 126850.7
688 128705.5
704 130593.6
720 132354.6
736 134413.2
752 136677.7
768 138015.2
784 140253.6
800 142683.2
816 144565.8
832 147995.3
848 149495.8
864 152164.3
880 155196.2
896 157772.3
912 158900.4
928 161569.9
944 164371.6
960 166952.1
976 168558.3
992 170707.5
1008 172417.2
1024 175282.7
1040 177102.9
1056 178897.8
1072 180230.8
1088 180933.3
1104 182313.6
1120 182871.5
1136 183936.2
1152 184547.4
1168 185207.9
1184 185556.1
1200 186113.3
1216 186468.7
1232 186711.4
1248 187119.4
1264 187580.1
1280 187877.5
1296 188385.3
1312 188629.3
1328 188903.3
1344 189124.3
1360 189251.9
1376 189424.5
1392 189831.2
1408 189972
1424 190291.4
1440 190563.3
1456 190891.2
1472 191111.9
1488 191273.5
1504 191424.4
1520 191577.4
1536 191761
1552 191956.2
1568 192091.6
1584 192340.3
1600 192484.8
1616 192574.8
1632 192674.6
1648 192821.6
1664 192915.4
1680 193009.1
1696 193110.9
1712 193235.6
1728 193322.9
1744 193426.4
1760 193493.8
1776 193575.5
1792 193718.6
1808 193821.6
1824 193974.6
1840 194065.4
1856 194135.8
1872 194236.4
1888 194357.4
1904 194587
1920 194662.7
1936 194795
1952 194943.3
1968 195047.5
1984 195117.6
2000 195211.5
};
\addlegendentry{CSQMI}
\addplot [semithick, black, forget plot]
table {%
0 200395
2000 200395
};
\addplot [semithick, black, dashed, forget plot]
table {%
0 180355.5
2000 180355.5
};
\end{axis}

\end{tikzpicture}